\documentclass{article}
\usepackage{ijcai25}

\usepackage{times}
\usepackage{soul}
\usepackage{url}
\usepackage[hidelinks]{hyperref}
\usepackage[utf8]{inputenc}
\usepackage[small]{caption}
\usepackage{graphicx}
\usepackage{amsmath}
\usepackage{amsthm}
\usepackage{booktabs}
\usepackage{algorithm}
\usepackage[switch]{lineno}
\usepackage{multirow}
\usepackage{colortbl}
\usepackage{xcolor}
\usepackage{amssymb}
\usepackage{algpseudocode}
\usepackage{verbatim}
\definecolor{lightblue}{RGB}{217, 237, 247}
\newcommand{\highlight}[1]{\colorbox{lightblue}{$#1$}}

\newcommand{\citet}[1]{\citeauthor{#1} \shortcite{#1}}

\newtheorem{theorem}{Theorem}
\newtheorem{assumption}{Assumption}

\title{Imitation Learning from Suboptimal Demonstrations via Meta-Learning An Action Ranker}

\author{
Jiangdong Fan$^1$ \and Hongcai He$^1$\and Paul Weng$^2$ \and Hui
Xu$^1$ \and Jie Shao$^{1,3}$\\
\affiliations
$^1$University of Electronic Science and Technology of China, Chengdu, China\\
$^2$Duke Kunshan University, Kunshan, China\\
$^3$Sichuan Artificial Intelligence Research Institute, Yibin, China\\
\emails \{jiangdongfan,hehongcai\}@std.uestc.edu.cn,
paul.weng@duke.edu, \{huixu.kim,shaojie\}@uestc.edu.cn}

\begin{document}

\maketitle

\begin{abstract}
A major bottleneck in imitation learning is the requirement of a
large number of expert demonstrations, which can be expensive or
inaccessible. Learning from supplementary demonstrations without
strict quality requirements has emerged as a powerful paradigm to
address this challenge. However, previous methods often fail to
fully utilize their potential by discarding non-expert data. Our key
insight is that even demonstrations that fall outside the expert
distribution but outperform the learned policy can enhance policy
performance. To utilize this potential, we propose a novel approach
named imitation learning via meta-learning an action ranker (ILMAR).
ILMAR implements weighted behavior cloning (weighted BC) on a
limited set of expert demonstrations along with supplementary
demonstrations. It utilizes the functional of the advantage function
to selectively integrate knowledge from the supplementary
demonstrations. To make more effective use of supplementary
demonstrations, we introduce meta-goal in ILMAR to optimize the
functional of the advantage function by explicitly minimizing the
distance between the current policy and the expert policy.
Comprehensive experiments using extensive tasks demonstrate that
ILMAR significantly outperforms previous methods in handling
suboptimal demonstrations. Code is available at
\url{https://github.com/F-GOD6/ILMAR}.
\end{abstract}

\section{Introduction}

Reinforcement learning has achieved notable success in various
domains, such as robot control
\cite{DBLP:journals/jmlr/LevineFDA16}, autonomous driving
\cite{DBLP:journals/tits/KiranSTMSYP22} and large-scale language
modeling \cite{DBLP:conf/icml/CartaRWLSO23}. However, its
application is significantly constrained by a carefully designed
reward function \cite{DBLP:conf/nips/Hadfield-Menell17} and the
extensive interactions with the environment
\cite{DBLP:journals/jmlr/GarciaF15}.

Imitation learning (IL) emerges as a promising paradigm to mitigate
these constraints. It derives high-quality policies from expert
demonstrations, thus circumventing the need for a predefined reward
function, often in offline settings where interaction with the
environment is unnecessary \cite{DBLP:journals/csur/HusseinGEJ17}.
However, to alleviate the compounding error issue---where errors
accumulate over multiple predictions, leading to significant
performance degradation---substantial quantities of expert
demonstrations are required \cite{DBLP:journals/jmlr/RossB10}.
Unfortunately, acquiring additional expert demonstrations is often
prohibitively expensive or impractical.

Compared with expert demonstrations, suboptimal demonstrations can
often be collected in large quantities at a lower cost. However, a
distributional shift exists between suboptimal and expert
demonstrations \cite{DBLP:conf/iclr/KimSLJHYK22}. Standard imitation
learning algorithms
\cite{DBLP:journals/neco/Pomerleau91,DBLP:conf/nips/HoE16}, which
process expert and non-expert demonstrations indiscriminately, may
inadvertently learn the deficiencies inherent in suboptimal
demonstrations, potentially degrading the quality of the learned
policies. Current approaches to addressing this issue often require
manual annotation of the demonstrations
\cite{DBLP:conf/icml/WuCBTS19} or interaction with the environment
\cite{DBLP:conf/nips/ZhangCSS21}, both of which are expensive and
time-consuming. Another category of methods, which has shown
considerable promise, utilizes a small-scale expert dataset along
with a large-scale supplementary dataset sampled from one or more
suboptimal policies
\cite{DBLP:conf/iclr/KimSLJHYK22,DBLP:conf/icml/XuZYQ22,DBLP:conf/nips/LiXQ0L23}.
This paper is focused on exploring this particular setup.

Previous studies often train a discriminator to distinguish between
expert and non-expert demonstrations, performing weighted imitation
learning on the supplementary dataset. However, during the training
of the discriminator, labeled expert demonstrations are assigned a
value of 1, while the demonstrations from the supplementary dataset
are assigned a value of 0 and discarded. Given the limited scale of
the labeled expert dataset, the supplementary dataset may contain a
substantial amount of unlabeled expert demonstrations, leading to a
positive-unlabeled classification problem
\cite{DBLP:conf/kdd/ElkanN08}. Furthermore, supplementary dataset
often includes many high-quality demonstrations that, although not
optimal, could improve policy performance when selectively
leveraged, especially in cases of insufficient expert demonstrations
\cite{DBLP:conf/icml/XuZYQ22}. Thus, these weighting imitation
learning methods based on the expert distribution tend to discard
high-quality non-expert demonstrations, failing to fully leverage
suboptimal demonstrations. Our key insight is that learning from
demonstrations outside the expert distribution, which still
outperform the learned policy, can further enhance policy
performance.

\begin{figure}[t]
    \centering
    \includegraphics[width=0.8\columnwidth]{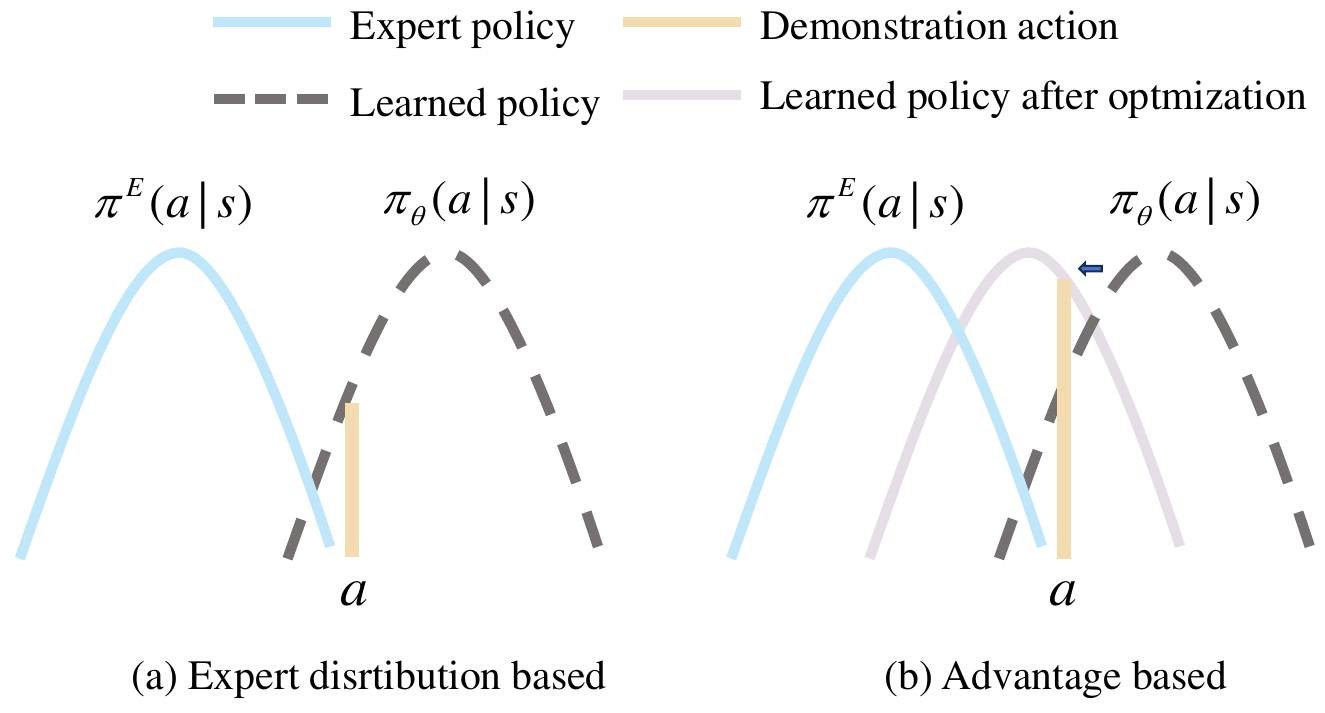}
    \caption{Left: Weighted imitation learning based on the expert distribution.
Right: Weighted imitation learning based on the advantage function.
Imitation learning weighted by the expert distribution fails to
update the policy when non-expert demonstrations exceed the learned
policy, wasting valuable data. Conversely, weighting by the
advantage function recognizes superior non-expert actions,
optimizing the policy and enhancing performance.}
    \label{fig:motivation_of_ILMAR}
\end{figure}

Leveraging this insight, we introduce an offline imitation learning
algorithm designed to effectively learn from suboptimal
demonstrations. Built upon the task of weighted behavior cloning
(BC) \cite{DBLP:journals/neco/Pomerleau91}, our approach involves
training a discriminator that can assess the quality of
demonstrations. By applying weights based on the advantage function
relative to the learned policy, we ensure that high-quality
demonstrations outside the expert distributions are retained, as
depicted in Figure~\ref{fig:motivation_of_ILMAR}. To further enhance
performance, we introduce meta-goal, a bi-level optimization
meta-learning framework aimed at improving weighted imitation
learning from suboptimal demonstrations. Meta-goal explicitly
minimizes the distance between the learned and the expert policies,
allowing the discriminator to optimally assign weights. This results
in a policy that closely emulates the expert policy. Consequently,
we have named our algorithm imitation learning from suboptimal
demonstrations via meta-learning an action ranker (ILMAR).

Our main contributions are as follows:
\begin{itemize}
    \item We propose ILMAR, a novel and high-performing algorithm based on
weighted behavior cloning in suboptimal demonstration imitation
learning.
    \item We introduce the meta-goal method, which significantly improves the
performance of imitation learning from suboptimal demonstrations
based on weighted behavior cloning.
    \item We conduct extensive experiments to empirically validate the
effectiveness of ILMAR. The results demonstrate that ILMAR achieves
performance competitive with or superior to state-of-the-art
imitation learning algorithms in tasks involving both expert and
additional demonstrations.
\end{itemize}

\section{Related Work}

\paragraph{Imitation Learning with Suboptimal Demonstrations}

In imitation learning, a large number of expert demonstrations are
typically required to minimize the negative effects of compounding
errors. However, obtaining expert demonstrations is often expensive
or even impractical in most cases. Therefore, researchers have
turned to using suboptimal demonstrations to enrich the dataset
\cite{DBLP:conf/nips/LiXQ0L23,DBLP:conf/iclr/KimSLJHYK22}.
Traditional imitation learning methods, such as behavioral cloning
(BC) \cite{DBLP:journals/neco/Pomerleau91} and generative
adversarial imitation learning (GAIL) \cite{DBLP:conf/nips/HoE16},
often treat all demonstrations uniformly, which can lead to
suboptimal performance.

To address this issue, BCND \cite{DBLP:conf/iclr/SasakiY21} employs
a two-step training process to weight the suboptimal demonstrations
using a pre-trained policy. However, this method performs poorly
when the proportion of expert demonstrations in the suboptimal
dataset is low. CAIL \cite{DBLP:conf/nips/ZhangCSS21} ranks
demonstrations by superiority and assigns different confidence
levels to suboptimal demonstrations, but this approach requires
extensive interaction with the environment. ILEED
\cite{DBLP:conf/icml/BeliaevSESP22} leverages demonstrator identity
information to estimate state-dependent expertise, weighting
different demonstrations accordingly. The most similar studies to
ours are DWBC \cite{DBLP:conf/icml/XuZYQ22}, DemoDICE
\cite{DBLP:conf/iclr/KimSLJHYK22} and ISW-BC
\cite{DBLP:conf/nips/LiXQ0L23}, which weight suboptimal
demonstrations by leveraging a small number of expert demonstrations
along with supplementary demonstrations. However, these methods
discard a substantial portion of high-quality suboptimal
demonstrations within the supplementary dataset by focusing solely
on distinguishing expert demonstrations from non-expert
demonstrations. Our method is based on the advantage function, which
assigns weights by comparing demonstrations with the learned policy.
This strategy can effectively leverage these high-quality,
non-expert data.

\paragraph{Imitation Learning with Meta-Learning}

Meta-imitation learning is an effective strategy to address the lack
of expert demonstrations
\cite{DBLP:conf/nips/DuanASHSSAZ17,DBLP:conf/corl/FinnYZAL17}. It
typically involves acquiring meta-knowledge from other tasks in a
multi-task setting, enabling rapid adaptation to the target task
\cite{DBLP:conf/icml/FinnAL17,DBLP:conf/corl/GaoJ022}. Although our
approach utilizes a meta-learning framework, it is designed for a
single-task setting and uses meta-goal to optimize the model. The
study most similar to ours is meta-gradient reinforcement learning
\cite{DBLP:conf/nips/XuHS18}, which uses meta-gradients to obtain
optimal hyperparameters. In our approach, the use of meta-goal
allows the model to assign appropriate weights, leading to the
development of a satisfactory policy.

\section{Problem Setting}

\subsection{Markov Decision Process}

We formulate the problem of learning from suboptimal demonstrations
as an episodic Markov decision process (MDP), defined by the tuple
$M = \langle S, A, P, R, H, p_0, \gamma \rangle$, where $S$ is the
state space, $A$ is the action space, $H$ is the episode length,
$p_0$ is the initial state distribution, $P$ is the transition
function such that $P_h(s_{t+1}|s_t, a_t)$ determines the transition
probability of transferring to state $s_{t+1}$ by executing action
$a_t$ in state $s_t$, $R: S \times A \rightarrow \mathbb{R}$ is the
reward function, and $\gamma$ is the discount factor. Although we
assume that the reward function is deterministic, as is typically
the case with reinforcement learning, we do not utilize any reward
information in our approach.

A policy $\pi$ in an MDP defines a probability distribution over
actions given a state. The state-action value function $Q^\pi(s,a)$
and the state value function $V^\pi(s)$ for a policy $\pi$ are
defined as $Q^{\pi}(s,a) =
\mathbb{E}\left[\sum_{t=1}^{H}\gamma^{t}R(s_{t},a_{t})|s_{1}=s,a_{1}=a\right]$
and $V^{\pi}(s) =
\mathbb{E}\left[\sum_{t=1}^{H}\gamma^{t}R(s_{t},a_{t})|s_{1}=s\right]$.
The advantage function $A^\pi$ is defined as $A^\pi(s, a) = Q^\pi(s,
a)-V ^\pi(s)$, which measures the relative benefit of taking action
$a$ in state $s$ compared with the average performance of the policy
$\pi$ in that state.

\subsection{IL with Supplementary Demonstrations}

In imitation learning, it is typically assumed that there exists an
optimal expert policy $\pi^E$, and the goal is to have the agent
make decisions by imitating this expert policy. To mitigate the
issue of compounding errors, substantial quantities of expert
demonstrations are typically required. A promising solution is to
supplement the dataset with suboptimal demonstrations.

We use the expert policy to collect an expert dataset $\mathcal
D^E=\{\tau_1,\cdots\,\tau_{N_E}\}$, consisting of $N_E$
trajectories. Each trajectory is a sequence of state-action pairs
$\tau=\{s_1,a_1.\cdots,s_H,a_H\}$. The supplementary dataset
$\mathcal D^S=\{\tau_1,\cdots\,\tau_{N_S}\} $ is collected using one
or more policies, where $N_S$ is the number of supplementary
trajectories. In general, there is no strict quality requirement for
the policies used to collect the supplementary dataset. These
trajectories may originate from a wide range of policies, spanning
from near-expert level to those performing almost randomly.
Therefore, it is crucial to develop algorithms that can effectively
discern useful demonstrations from the supplementary dataset of
varying quality to learn a good policy. We combine the expert
dataset $\mathcal D^E$ with the supplementary datasets $\mathcal
D^S$ to form the full dataset $\mathcal D$.

Weighted behavior cloning is a classical approach to tackle this
challenge. It seeks to assign weights that reflect the expert level
of the demonstrations and then perform imitation learning on the
reweighted dataset. The optimization objective of weighted behavior
cloning is as follows:
\begin{equation}
     \min _{\pi} \underset{(s, a) \sim \mathcal{D}}{\mathbb{E}}[-  w(s, a) \log \pi(a|s)],
\end{equation}
where $w(s,a)$ is an arbitrary weight function, and $s$ and $a$ are
the state and action in demonstrations. When $w(s,a)=1$ for all $(s,
a) \in \mathcal{D}$, weighted behavior cloning degenerates to
vanilla BC. If $w(s,a)$ is the weight assigned by a discriminator
that distinguishes expert demonstrations, this objective aligns with
the optimization goal of ISW-BC \cite{DBLP:conf/nips/LiXQ0L23}. DWBC
\cite{DBLP:conf/icml/XuZYQ22} expands this by incorporating the
policy into discriminator training. The primary goal of weighted BC
is to filter out low-quality demonstrations and selectively learn
from valuable suboptimal demonstrations in the supplementary
dataset.

\section{Method}

In this section, we present a novel offline imitation learning
algorithm called imitation learning from suboptimal demonstrations
via meta-learning an action ranker. Our goal is to train a
discriminator capable of evaluating the relative benefit of the
learned policy and the demonstration policy, allowing us to fully
leverage suboptimal demonstrations in the supplementary dataset to
improve the learned policy. We propose a bi-level optimization
framework to enhance the performance of weighted behavior cloning
method, enabling the discriminator to automatically learn how to
assign appropriate weights to the demonstrations in order to achieve
high-performance policies. We also provide an explanation of the
weights assigned by our discriminator, which offers an intuitive
understanding of why our approach is effective.

\begin{figure}[t]
    \centering
    \includegraphics[width=1.0\columnwidth]{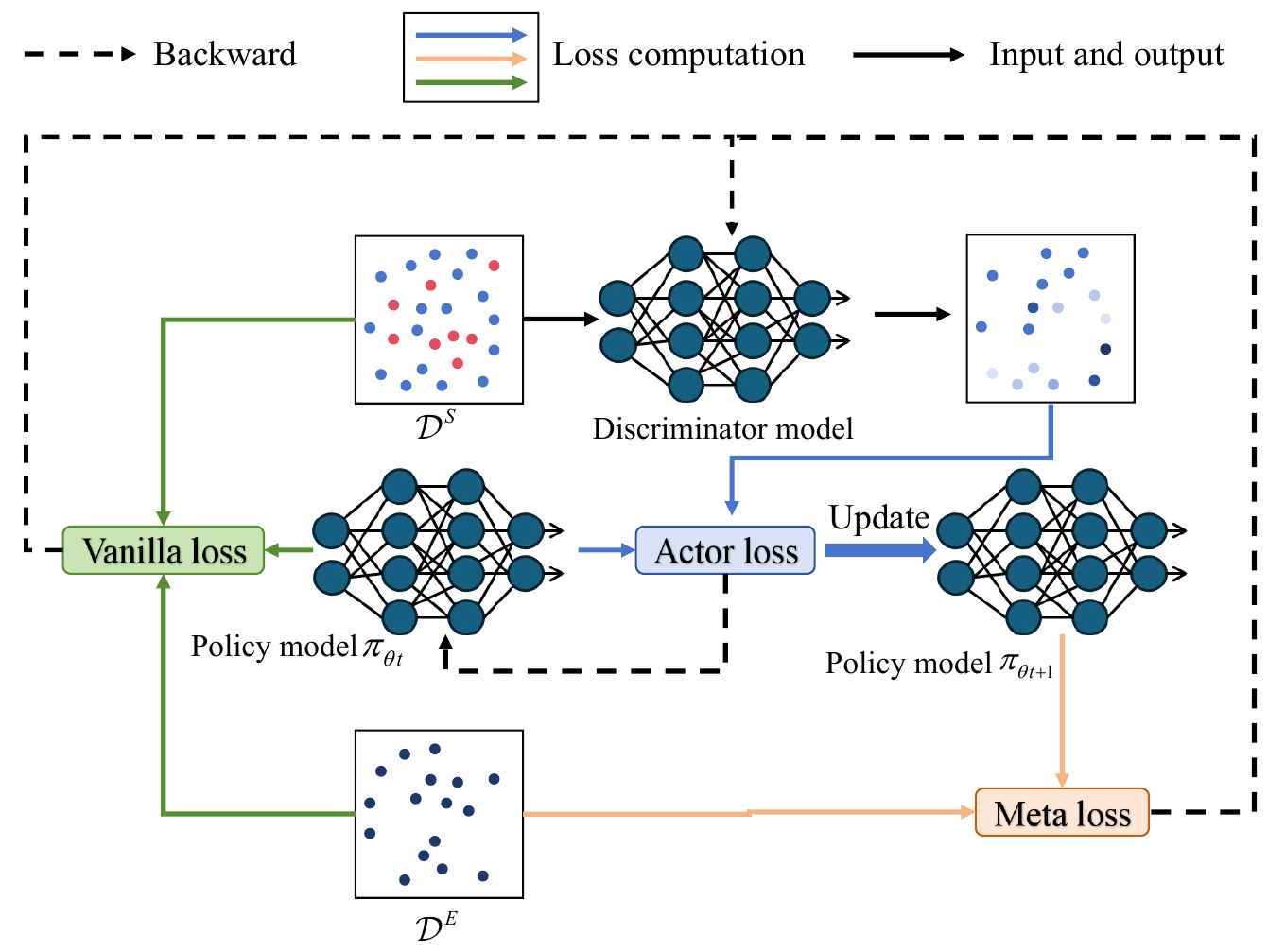}
    \caption{Illustration of our proposed ILMAR framework. Blue dots represent
demonstrations outperforming the learned policy, while red dots
represent those underperforming it. The intensity of blue indicates
the weight assigned by the discriminator. The discriminator
reweights the supplementary dataset $\mathcal D^S$, filtering out
inferior demonstrations. The policy is then trained on the
reweighted demonstrations via behavioral cloning. After updates, the
gap between the updated policy and expert demonstrations in
$\mathcal D^E$ is used to compute the meta loss. Additionally, the
relative performance of expert demonstrations, suboptimal
demonstrations, random policy, and the current policy determine the
vanilla loss. The discriminator is updated based on the meta loss
and the vanilla loss to improve weighting.}
    \label{fig:framework}
\end{figure}

\subsection{Imitation Learning via Learning An Action Ranker}

In our framework, we utilize a model parameterized by $\theta$ as
the policy $\pi$ and another model parameterized by $\psi$ as the
discriminator $C$, which evaluates the relative benefit of
demonstrations compared with the learned policy.

It is evident that we can avoid learning low-quality demonstrations
if we have access to an advantage function. The optimization
objective of policy $\pi$ can be written as:
\begin{equation}
\min _{\pi} \underset{(s, a) \sim \mathcal{D}}{\mathbb{E}}[-  A^\pi(s, a) \log \pi(a|s)],
\end{equation}
which is widely employed in reinforcement learning to guide policy
updates. However, directly learning an advantage function without
reward information and without interacting with the environment is
challenging. To implement an equivalent form of the advantage
function, we introduce a weight function
$w(s,a,\pi)=\mathbb{I}(A^\pi(s, a)>0)$, where $\mathbb{I}$ is an
indicator function that assigns a value of 1 when the condition
$(A^\pi(s, a)>0$ holds, and 0 otherwise.

This formulation ensures that behavior cloning is performed
selectively, focusing only on demonstrations that outperform the
learned policy. By doing so, the modification aligns the policy
update process with the use of the advantage function, maintaining
consistency in prioritizing superior demonstration actions.
Importantly, this approach retains the core effect of leveraging the
advantage function for guiding policy updates without altering its
fundamental role in policy optimization.

In practice, we use $\mathbb{I}(\mathbb{P}(A^\pi(s, a))>\frac{1}{2})
\cdot (\mathbb{P}(A^\pi(s, a)>0))$ to approximate
$\mathbb{I}(A^\pi(s, a)>0)$. The output of the discriminator
corresponds to the probability that $A^\pi(s, a)>0$. We then define
the weight function as $w(s,a,\pi)=\mathbb{I}(\mathbb{P}(A^\pi(s,
a))>\frac{1}{2}) \cdot (\mathbb{P}(A^\pi(s, a)>0))$, meaning that
when the demonstration action is more likely to outperform the
learned policy, we increase the probability of sampling that action.

We train the discriminator using the available information as
follows: when the demonstrations come from expert demonstrations,
the discriminator outputs $\mathbb{P}(A^\pi(s, a)>0)=1$, for all
$(s,a) \in D^E$, indicating that the expert policy outperforms the
learned policy. Furthermore, we introduce random actions into the
action space, which can be viewed as actions chosen by a random
policy $\pi^r$. In this case, we have $\mathbb{P}(A^{\pi^r}(s,
a)>0)=1$, where $s\in D,a \in \mathcal{D}$ or $\pi_\theta$,
indicating that all the expert, suboptimal and the learned policies
outperform the random policy. Since it is difficult to obtain
actions from both the expert and suboptimal policies in the same
state, we do not directly update the discriminator based on a direct
comparison between expert and suboptimal policies. Thus, the
training objective of the discriminator is given by:
\begin{equation}
\label{eq:vanilla loss}
    \min_C       \mathbb{E}_{s,  a_1, a_2 }
    [\log(C(s,a_1,a_2)]+[\log(1-C(s,a_2,a_1)],
\end{equation}
where $s \sim \mathcal D$ and $a_1, a_2$ are sampled from either
$D$, $\pi_\theta(s)$ or $\pi^r(s)$). Here, $a_1$ is not inferior to
$a_2$. For simplicity, $\pi_\theta(s)$ is written in a deterministic
form. If the policy is stochastic, the action can be sampled from
the policy distribution or taken as the expectation of the action
distribution as input to the discriminator. In implementation, our
discriminator determines the relative advantage between two actions.
Therefore, we name our algorithm imitation learning via learning an
action ranker.

The weight function $w(s,a,\pi)$ is a functional of the advantage
function. To clarify why weighting based on the advantage function
should outperform weighting based on the expert distribution, let us
consider a scenario where the supplementary dataset consists
entirely of suboptimal demonstrations (high-quality but non-expert),
and both the advantage-based and expert-distribution-based
discriminators are well-trained. During the initial training phase,
all demonstrations in the supplementary dataset outperform the
learned policy. The advantage-based weighting approach would assign
large weights to all these suboptimal demonstrations, effectively
utilizing the entire supplementary dataset. In contrast, the
expert-distribution-based weighting method assigns small weights to
all demonstrations (since it lies outside the expert distribution),
resulting in the policy learning primarily from a limited set of
expert demonstrations and thus wasting a significant number of
high-quality suboptimal demonstrations.

\paragraph{Algorithmic Update Procedure}

We now detail the update process at time step $t$. Let the policy
$\pi_{\theta_t}$ be parameterized by $\theta_t$. We employ the
weights $w_{\psi_t}(s, a, \pi_{\theta_t}) = \mathbb{I}(C(s, a,
\pi_{\theta_t}))>\frac{1}{2}) \cdot C(s, a, \pi_{\theta_t})$
obtained from the discriminator $C_{\psi_t}$ for weighted behavior
cloning. Here, $s$ and $a$ represent the state and action in the
demonstration. Using maximum likelihood estimation, we define the
actor loss $L_{actor}$ as:
\begin{equation}
    L_{actor}=- \frac{1}{|\mathcal D|}\sum _{(s,a)\in \mathcal D}w(s,a,\pi_{\theta_t}) \log \pi_{\theta_t}(a|s).
\label{eq:policy loss}
\end{equation}
Accordingly, we update $\theta_t$ as follows:
\begin{equation}
    \theta_{t+1}=\theta_t - \mu \nabla_\theta L_{actor}(s,a;\theta_t,\psi_{t}),
    \label{eq:update theta}
\end{equation}
where $\mu$ is the learning rate of policy. We then update the
discriminator $C$ by minimizing the objective in
Eq.~\eqref{eq:vanilla loss}.

\subsection{Meta-Goal for Weighted Behavior Cloning}

To enhance weighted behavior cloning for learning from suboptimal
datasets, we propose the meta-goal approach. In weighted behavior
cloning, the discriminator should assign weights to the suboptimal
dataset such that the resulting policy closely resembles the expert
policy, effectively recovering the expert distribution. This goal
can be instantiated using the Kullback-Leibler (KL) divergence:
\begin{equation}
\min _{\pi,C} D_{\mathrm{KL}}\left(\pi^E \| \pi\right),
\end{equation}
where $D_{\mathrm{KL}}\left(\pi^E \| \pi\right)=\mathbb{E}_{s \sim
d^{\pi^E}}\left[D_{\mathrm{KL}}\left(\pi^E(\cdot \mid s) \|
\pi(\cdot \mid s)\right)\right]$ and $d^{\pi^E}$ is the stationary
distribution of the expert policy. We cannot directly optimize the
discriminator using this objective. Inspired by the meta-gradient
methods \cite{DBLP:conf/nips/XuHS18}, we adopt a bi-level
optimization framework. Specifically, drawing on the ideas of
expectation-maximization (EM), we proceed as follows: in the inner
optimization loop, we perform weighted behavior cloning using the
current discriminator to update the policy; in the outer
optimization loop, we then adjust the discriminator parameters based
on the resulting difference between the learned policy and the
expert policy. Through this nested optimization process, the
meta-goal method effectively guides the discriminator to assign
weights that lead to a policy closely resembling the expert.

\paragraph{Algorithmic Update Procedure}

We now detail the update process at time step $t$. First, we
optimize the policy using Eq.~\eqref{eq:policy loss}. Notably,
during this update, we retain the gradient of $w_{\psi_t}(s, a,
\pi_{\theta_t})$ with respect to $\psi_t$. This allows us to
incorporate these gradients into the outer optimization loop that
follows. We estimate the discrepancy between the learned policy and
the expert policy using demonstrations from the expert dataset
$D^E$. Specifically, we define the loss of meta-goal (i.e., meta
loss) as:
\begin{equation}
    L_{meta}=- \frac{1}{|\mathcal D^E|}\sum _{\substack{(s,a)\in \mathcal D^E}}\log
    \pi_{\theta_{t+1}}(a|s).
    \label{eq:meta_loss}
\end{equation}
We then update $\psi_t$ as follows:
\begin{equation}
    \psi_{t+1} = \psi_t - \varphi \nabla_\psi L_{meta}(s,a;\psi_t,\theta_{t+1}),
\end{equation}
where $\varphi$ is the learning rate for the discriminator
parameters. By applying the chain rule, the gradient $\frac{\partial
L_{meta}}{\partial \psi}$ can be expressed as:
\begin{equation}
     \mu \frac{1}{|\mathcal D|} \frac{\partial L_{meta}}{\partial \theta_{t+1}}  \sum _{(s,a)\in \mathcal D} \frac {\partial^2 w(s,a,\pi_{\theta_t}) \log \pi_{\theta_t}(a|s)}{\partial \psi \partial \theta_t},
\label{eq:chain rule}
\end{equation}
where $\mu$ is the policy learning rate (as defined in
Eq.~\eqref{eq:update theta}). We provide the detailed derivation of
Eq.~\eqref{eq:chain rule} in the supplementary material. In
practice, we combine the meta-goal approach with the original
weighted behavior cloning framework. We refer to the original
discriminator update objective as the vanilla loss. Thus, the final
discriminator loss function $L_C$ is a composite of the meta loss
$L_\text{meta}$ and the vanilla loss $L_\text{vanilla}$:
\begin{equation}
L_C = \alpha L_\text{meta} + \beta L_\text{vanilla},
\label{eq:total_loss}
\end{equation}
where $\alpha$ and $\beta$ are hyperparameters controlling the
relative influence of the meta loss and the vanilla loss on the
updates of the discriminator. This composite formulation enables the
discriminator to leverage both the learned policy-expert discrepancy
and the original objective, ultimately improving performance in
handling suboptimal demonstrations.

By integrating meta-goal with imitation learning via learning an
action ranker, we derive our complete method, imitation learning
from suboptimal demonstrations via meta-learning an action ranker
(ILMAR). The complete algorithm is formally presented as
Algorithm~\ref{alg:ILMAR}.

\begin{algorithm}[t]
\caption{ILMAR} \label{alg:ILMAR}
\begin{algorithmic}[1]
\State \textbf{Input:} Expert demonstration dataset $\mathcal D^E$
and the full dataset $\mathcal D$, policy learning rate $\mu$,
discriminator learning rate $\varphi$ \State Initialize policy
parameter $\theta_0$ and discriminator parameter $\psi_0$ \For{$
t=0,1,\cdots$ }
    \State Sample $N_1$ state-action pairs $d^E$ from $\mathcal D^E$ and $N_2$ state-action pairs $d$ from $\mathcal D$
    \vskip 0.5em
    \State Update $\theta_t$ according to Eq.~\eqref{eq:update theta}
    \State Update $\psi_t$ with $\nabla_\psi L_{C}$ (see Eqs.~\eqref{eq:vanilla loss}, \eqref{eq:meta_loss}, \eqref{eq:total_loss})
\EndFor
\end{algorithmic}
\end{algorithm}

\subsection{Theoretical Results}

When updating the discriminator network with meta-goal, ILMAR
employs a bi-level optimization framework, where the policy network
is updated in the inner loop and the discriminator network is
updated in the outer loop. The convergence results for similar
bi-level optimization problem are established in prior work
\cite{DBLP:conf/nips/ZhangCSS21}. Based on these results, we analyze
the convergence properties of the proposed discriminator and
demonstrate why we recommend applying meta-goal to enhance the
original weighted behavior cloning method, rather than using it as
an independent approach.

We introduce the following assumption.
\begin{assumption}[Lipschitz Smooth Function Approximators]
\label{ass:Lipschitz smooth} The discriminator loss function
$L_{\text{C}}$ is Lipschitz-smooth with constant $L$. Specifically,
for any parameters $\theta_0$ and $\theta_1$, the following
condition holds:
\[
\|\nabla L_{\text{C}}(\theta_1) - \nabla L_{\text{C}}(\theta_2)\| \leq L \|\theta_1 - \theta_2\|.
\]
\end{assumption}
Assumption~\ref{ass:Lipschitz smooth} stipulates that $L_{\text{C}}$
possesses Lipschitz smoothness and its first and second-order
gradients are bounded. This condition is satisfied when the trained
policy is Lipschitz-continuous and differentiable and the
discriminator output is clipped with a small constant $\epsilon>0$
to the range $[0+\epsilon,1-\epsilon]$. Such an assumption is
considered mild and is commonly adopted in numerous studies
\cite{DBLP:conf/nips/VirmauxS18,DBLP:conf/iclr/MiyatoKKY18,DBLP:conf/siggraph/LiuWJFL22,DBLP:conf/icml/XuZYQ22}.
Under this assumption, we have:
\begin{theorem}
\label{th:convergence} The discriminator loss decreases
monotonically (i.e., $L_{C}(\theta_{t+1} ) \leq L_{C}(\theta_{t})$)
under the condition that there exists a constant $K>0$ such that the
following inequality holds:
$\nabla_\theta\mathcal{L}_{C}(\theta_{t+1}
)^\top\nabla_\theta\mathcal{L}_{actor}(\theta_t,\psi_{t})\geq
K||\nabla_\theta\mathcal{L}_{actor}(\theta_t,\psi_{t})||^2$, and the
learning rate satisfies $\mu \leq \frac{2K}{L}$.
\end{theorem}
Theorem~\ref{th:convergence} demonstrates that, under the given
inequality, the policy updates ensure a monotonic decrease in the
discriminator loss. However, this inequality assumes that the
gradient directions of $L_C$ and $L_{actor}$ are closely aligned
\cite{DBLP:conf/nips/ZhangCSS21}. When the discriminator is updated
solely using the meta loss, it may become trapped in local optima or
fail to converge, as meta-goal, based on an EM-like approach,
assigns weights tentatively and lacks the guidance of prior
knowledge.

To address this issue, we incorporate prior knowledge by
simultaneously updating the discriminator using a vanilla loss. The
manually designed vanilla loss constrains the update direction of
the discriminator, ensuring that the gradient directions of $L_C$
and $L_{actor}$ remain closely aligned. This adjustment not only
corrects but also stabilizes the updates of the discriminator,
enhancing convergence. The experimental results strongly support
this theoretical claim, as shown in the corresponding sections.

\section{Experiments}

In this section, we conduct experiments to evaluate and understand
ILMAR. Specifically, we aim to address the following questions:
\begin{enumerate}
\item How does ILMAR perform compared with previous suboptimal
demonstrations imitation learning algorithms?
\item How does the proportion of suboptimal demonstrations affect the
performance of ILMAR?
\item Is meta-goal compatible with other algorithms, and does it enhance
their performance?
\end{enumerate}

\subsection{Comparative Evaluations}
\label{sec:compare_ILMAR}

\paragraph{Datasets}

In this section, we evaluate the effectiveness of ILMAR by
conducting experiments on the MuJoCo continuous control environments
\cite{DBLP:conf/iros/TodorovET12} using the OpenAI Gymnasium
framework \cite{towers_gymnasium}. We conduct experiments on four
MuJoCo environments: Ant-v2, Hopper-v2, Humanoid-v2 and Walker2d-v2.
We collect expert demonstrations and additional suboptimal
demonstrations and conduct the evaluation as follows. For each
MuJoCo environment, we follow prior dataset collection methods
\cite{DBLP:conf/icml/WuCBTS19,DBLP:conf/nips/ZhangCSS21}. Expert
agents are trained using PPO
\cite{DBLP:journals/corr/SchulmanWDRK17} for Ant-v2, Hopper-v2 and
Walker2d-v2, and SAC \cite{DBLP:journals/corr/abs-1812-05905} for
Humanoid-v2. We consider three tasks (T1, T2, T3) with a shared
expert dataset $\mathcal D^E$ consisting of one expert trajectory.
The supplementary datasets include expert and suboptimal
trajectories at ratios of 1:0.25 (T1), 1:1 (T2), and 1:4 (T3). Each
supplementary suboptimal dataset $\mathcal D^S$ contains 400 expert
trajectories and 100, 400, or 1600 suboptimal trajectories generated
by four intermediate policies sampled during training, with equal
contributions from each policy. We visualized the distributional
differences between the additional datasets and the expert
demonstrations under three task settings on Ant-v2. The detailed
methodology for the visualization is provided in the supplementary
materials. As shown in Figure~\ref{fig:task_setting}, the
distributional differences between expert and supplementary datasets
increase as the proportion of expert demonstrations decreases,
challenging the algorithm with more suboptimal data.

\begin{figure}[t]
    \centering
    \includegraphics[width=1.0\linewidth]{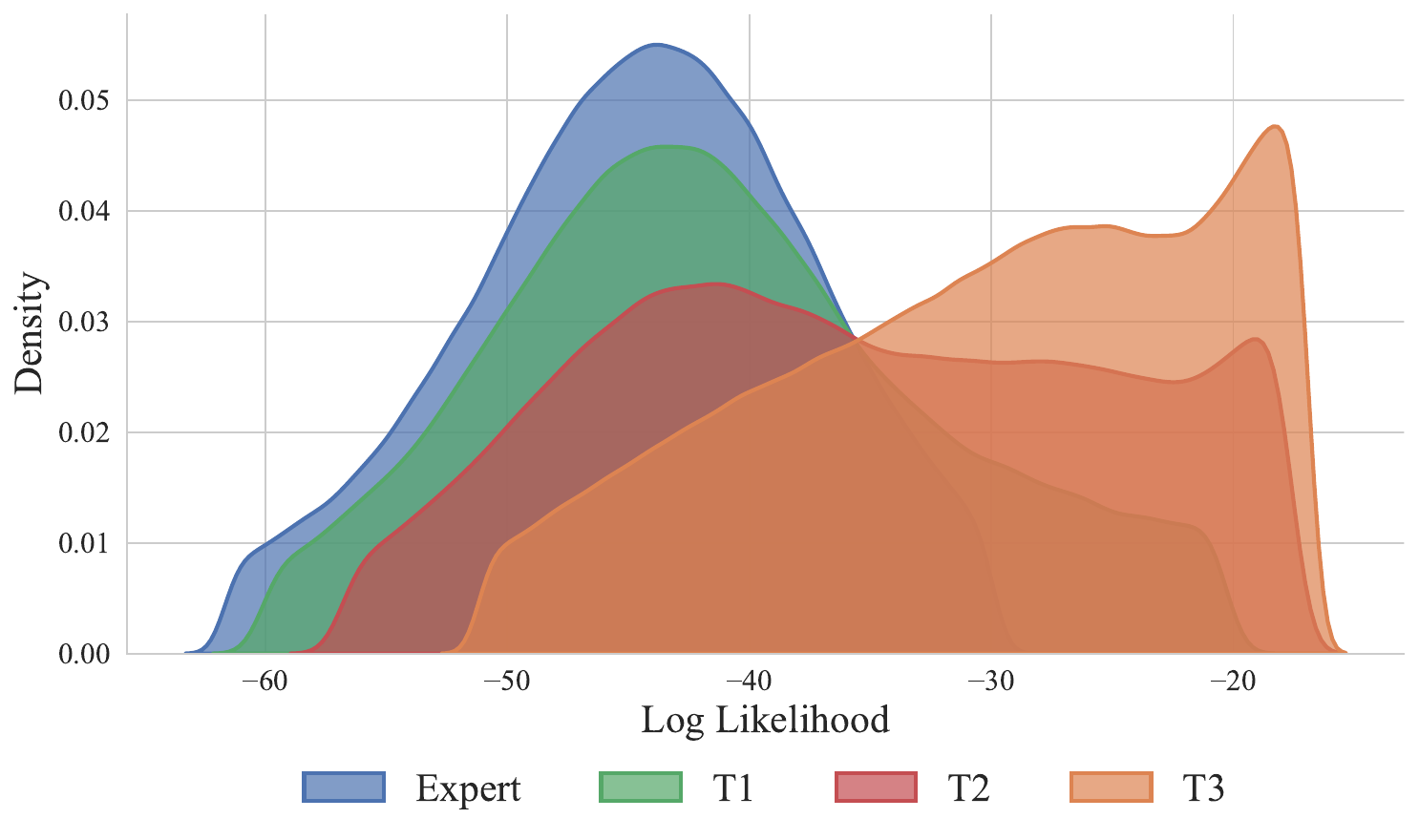}
    \caption{The kernel density estimates (KDE) of the log-likelihood for the
expert demonstrations and the suboptimal datasets on Ant-v2 of T1,
T2, and T3, based on a variational autoencoder (VAE) model.}
    \label{fig:task_setting}
\end{figure}

\paragraph{Baselines}

We compare ILMAR with five strong baseline methods in our problem
setting including BC, BCND \cite{DBLP:conf/iclr/SasakiY21}, DemoDICE
\cite{DBLP:conf/iclr/KimSLJHYK22}, DWBC
\cite{DBLP:conf/icml/XuZYQ22}, and ISW-BC
\cite{DBLP:conf/nips/LiXQ0L23}. For these methods, we use the
hyperparameter settings specified in their publications or in their
codes. The training process is carried out for 1 million
optimization steps. We evaluate the performance every 10,000 steps
with 10 episodes. More experimental details are provided in the
supplementary material.

\paragraph{Results}

\begin{figure}[t]
    \centering
    \includegraphics[width=1.0\linewidth]{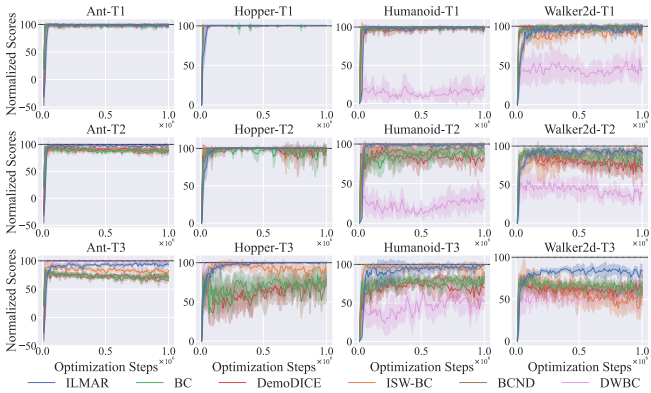}
    \caption{Training curves of ILMAR and baseline algorithms on tasks T1, T2,
T3. The y-axis represents the normalized scores of the algorithm
during training. The solid line corresponds to the average
performance under five random seeds, and the shaded area corresponds
to the 95\% confidence interval.}
    \label{fig:baseline}
\end{figure}

Table~\ref{tab:compare} and Figure~\ref{fig:baseline} present the
performance of different methods on the four environments under
three task settings. In Table~\ref{tab:compare}, the values are the
mean reward over the final 5 evaluations and 5 seeds, with
subscripts indicating standard deviations. Score in the table is the
average normalized score across environments. The normalized score
in one environment is computed as follows:
\[
    \text{score}= 100 \times \frac{\text{mean reward - random reward}}{\text{expert reward} -  \text{random reward}}.
\]

\begin{table}[t]
    \centering
    \resizebox{1.0\linewidth}{!}{
    \begin{tabular}{ccccccc}
        \toprule
        Task setting & Environment & Ant & Hopper & Humanoid & Walker2d  & Score \\
        \midrule
        & Random  & -75  & 15   & 122 & 1   & 0\\
        & Expert  & 4761  & 3635   & 7025 & 4021   & 100\\
        \midrule
        \multirow{6}{*}{\textbf{T1}}
        & BC  & 4649$_{\pm 69}$    & $3650{\pm 3}$  & $\highlight{6959{\pm 72}}$  & $3882{\pm 92}$ & 98.42\\
        & BCND  &$4694_{\pm 27}$ &   $3651_{\pm 2}$      & $6957_{\pm 47}$  &$\highlight{3961_{\pm 107}}$ & \highlight{99.15}\\
        & DemoDICE   &$4672_{\pm 61}$ &   $3651_{\pm 1}$      & $6917_{\pm 83}$  &$\highlight{3952_{\pm 116}}$ & 98.83\\
        & DWBC   &$\highlight{4718_{\pm 39}}$ &   $3650_{\pm 1}$      & $1337_{\pm 662}$  &$1818_{\pm 467}$ & 65.58\\
        & ISW-BC    &$\highlight{4712_{\pm 20}}$  &   $\highlight{3652_{\pm 1}}$  &$6958_{\pm 82}$ &$3672_{\pm 104}$ & 97.45 \\
        & ILMAR (ours) &$4689_{\pm 74}$   &$\highlight{3652_{\pm 2}}$       & $\highlight{6964_{\pm {81}}}$  & $3934_{\pm97}$ & $\highlight{98.98}$  \\
        \midrule
        \multirow{6}{*}{\textbf{T2}}
        & BC  & 4222$_{\pm82}$    & $3461_{\pm 335}$  & $6266_{\pm 345}$  & $3378_{\pm 108}$ & 89.26\\
        & BCND  & 4099$_{\pm60}$    & $3606_{\pm 46}$  & $6670_{\pm 382}$  & $3061_{\pm 108}$ & 89.12\\
        & DemoDICE   &$4192_{\pm 92}$ &   $3527_{\pm 183}$      & $5719_{\pm 791}$  &$2880_{\pm 462}$ & 84.49\\
        & DWBC   &$\highlight{4725_{\pm 52}}$ &   $\highlight{3643_{\pm 7}}$      & $2273_{\pm 1096}$  &$1574_{\pm 156}$ & 67.45\\
        & ISW-BC    &$4425_{\pm 54}$  &   $3583_{\pm 114}$  &$\highlight{6977_{\pm 63}}$ &$\highlight{3802_{\pm 78}}$ & \highlight{96.37}  \\
        & ILMAR (ours) &$\highlight{4654_{\pm 67}}$   &$\highlight{3651_{\pm 0}}$       & $\highlight{6958_{\pm 38}}$  & $\highlight{3728_{\pm {162}}}$ & $\highlight{97.49}$  \\
        \midrule
        \multirow{6}{*}{\textbf{T3}}
        & BC  & 3411$_{\pm166}$    & $2704_{\pm 388}$  & $5420_{\pm 205}$  & $2454_{\pm 267}$ & 71.04\\
        & BCND  & 3149$_{\pm56}$    & $2336_{\pm 599}$  & $5683_{\pm 296}$  & $2529_{\pm 215}$ & 68.56\\
        & DemoDICE   &$3462_{\pm 117}$ &   $2736_{\pm 189}$      & $4808_{\pm 890}$  &$\highlight{2564_{\pm 259}}$ & 69.99\\
        & DWBC   &$\highlight{4674_{\pm 59}}$ &   $\highlight{3649_{\pm 2}}$      & $3671_{\pm 622}$  &$2120_{\pm 329}$ & 75.68\\
        & ISW-BC    &$3770_{\pm 57}$  &   $3236_{\pm 422}$  &$\highlight{6534_{\pm 586}}$ &$1926_{\pm 522}$ & \highlight{77.32}  \\
        & ILMAR (ours) &$\highlight{4419_{\pm 67}}$   &$\highlight{3551_{\pm 106}}$       & $\highlight{6909_{\pm {46}}}$  &$\highlight{3259_{\pm {226}}}$ & $\highlight{93.16}$  \\
        \bottomrule
    \end{tabular}}
    \caption{Performance of ILMAR and baseline algorithms on Ant-v2, Hopper-v2,
Walker2d-v2 and Humanoid-v2 over the final 5 evaluations and 5
seeds. The best two results are highlighted. ILMAR significantly
outperforms existing imitation learning methods from suboptimal
datasets.}
    \label{tab:compare}
\end{table}

The results demonstrate that ILMAR achieves performance that is
competitive with or superior to state-of-the-art IL algorithms in
tasks involving expert and additional suboptimal demonstrations
across three task-setting. Additionally, we can observe that when
the proportion of suboptimal demonstrations in the additional
dataset is low, the distributional differences between the
supplementary and expert datasets are minimal. Under such
conditions, all algorithms, even BC, that do not differentiate
between the supplementary and expert datasets, can achieve
performance close to that of the expert. However, as the proportion
of suboptimal demonstrations increases, it becomes essential to
design sophisticated algorithms to address the challenges posed by
suboptimal demonstrations. With an increasing proportion of
suboptimal demonstrations, the advantages of ILMAR over other
methods become increasingly pronounced.

\subsection{Ablation Studies}

In this section, we conduct ablation studies to analyze the effects
of the different components of the loss function.

Table~\ref{tab:abl_loss} presents the results for ILMAR when
evaluated using only the naive loss or only the meta loss across the
MuJoCo experiments. First, ILMAR outperforms the best baseline even
without using meta-goal. This validates that weighting suboptimal
demonstrations based on advantage function can enhance performance.
When ILMAR is updated solely using meta-goal, its performance is
slightly worse than that of BC. This phenomenon occurs because
training the discriminator with only the meta-goal optimization
leads to instability, making it difficult for the model to converge
correctly. Further analyses and experiments have been conducted to
understand this issue in greater depth, as detailed in the
supplementary materials.

\begin{table}[t]
    \centering
    \resizebox{1.0\linewidth}{!}{
    \begin{tabular}{c|cccc|c}
        \toprule
        Environment & Ant & Hopper & Humanoid & Walker2d  & Score \\
        \midrule
        ILMAR (vanilla loss)  & 4232$_{\pm 90}$    & $3045_{\pm 451}$  & $6872_{\pm 114}$  &$3236_{\pm 268}$ &87.75\\
        ILMAR (meta loss)   &$3260_{\pm 443}$ &   $2662_{\pm 1140}$      & $2443_{\pm 1959}$  &$2357_{\pm 998}$ &58.58\\
        ILMAR &$\boldsymbol{4419_{\pm 67}}$   &$\boldsymbol{3551_{\pm 106}}$       & $\boldsymbol{6909_{\pm \boldsymbol{46}}}$  &$\boldsymbol{3259_{\pm \boldsymbol{226}}}$ & $\boldsymbol{93.16}$  \\
        \bottomrule
    \end{tabular}}
    \caption{Performance results of ILMAR using only naive loss and meta loss for
updates on the MuJoCo environments. The best results are in bold.}
    \label{tab:abl_loss}
\end{table}

When ILMAR optimizes with both the vanilla loss and the meta loss,
its performance improves further. This confirms that meta-goal
effectively enhances performance of weighted behavior cloning. As
indicated in our theoretical analysis, the vanilla loss provides
prior knowledge that constrains the parameter update direction,
facilitating the convergence of the meta loss and enhancing the
performance of model.

\subsection{Meta-goal Used in Other Algorithms}

In this section, we apply meta-goal to ISW-BC and DemoDICE to
evaluate its compatibility with other algorithms. Specifically, we
update the discriminator of ISW-BC and DemoDICE using both the meta
loss and the discriminator loss of original algorithm, as shown in
Eq.~\eqref{eq:total_loss}. Table~\ref{tab:meta-goal} presents the
results of employing meta-goal in ISW-BC and DemoDICE across the
MuJoCo experiments. We observe that incorporating meta-goal leads to
significant performance improvements for both DemoDICE and ISW-BC in
nearly all environments. This further validates the effectiveness
and robustness of meta-goal, demonstrating its compatibility with
other algorithms.

\begin{table}[t]
    \centering
    \resizebox{1.0\linewidth}{!}{
    \begin{tabular}{cc|cccc}
        \toprule
        Environment && Ant & Hopper & Humanoid & Walker2d   \\
        \midrule
        \multirow{2}{*}{DemoDICE}   &&$3462_{\pm 117}$ &   $2736_{\pm 189}$      & $4808_{\pm 890}$  &$2564_{\pm 259}$ \\
        &+Meta-goal&$\mathbf{3654_{\pm 71}}$  &   $\mathbf{2907_{\pm 140}}$  &$\mathbf{5252_{\pm 448}}$ &$\mathbf{2811_{\pm 145}}$  \\
        \midrule
        \multirow{2}{*}{ISW-BC}   &  &$3770_{\pm 57}$  &   $3236_{\pm 422}$  &$6534_{\pm 586}$ &$1926_{\pm 522}$\\
         &+Meta-goal    &$\mathbf{3974_{\pm 151}}$  &   $\mathbf{3294_{\pm 143}}$  &$\mathbf{6907_{\pm 87}}$ &$\mathbf{3186_{\pm 185}}$  \\
        \bottomrule
    \end{tabular}}
     \caption{Performance of other algorithms using meta-goal on the MuJoCo
environments. The best results are in bold.}
    \label{tab:meta-goal}
\end{table}

\subsection{Additional Analysis of Results}

To further understand ILMAR, we explore the relationship between the
weights learned by ILMAR and the actual reward values. The results
indicate a clear monotonic positive correlation between the weights
assigned by ILMAR and the true rewards. Specifically, we calculate
the Spearman rank correlation coefficient, which measures the
strength and direction of the monotonic relationship between two
variables, between the weights and true rewards in the task setting
T3 for Ant-v2 and Humanoid-v2. The formula is given by:
\[
\rho = 1 - \frac{6 \sum d_i^2}{n(n^2 - 1)}
\]
where:
\begin{itemize}
    \item \( d_i = \text{rank}(x_i) - \text{rank}(y_i) \) is the difference between the ranks of corresponding values \( x_i \) and \( y_i \) from the two variables.
    \item \( n \) is the number of observations.
\end{itemize}
This coefficient provides a robust measure of the monotonic
relationship, making it suitable for evaluating non-linear
dependencies in the data. The Spearman rank correlation coefficients
between the weights and true rewards for Ant-v2 and Humanoid-v2 are
0.7862 and 0.7220, respectively. These results strongly validate
that the weights assigned by ILMAR effectively represent the
superiority of demonstration actions, enabling the model to learn
from suboptimal demonstrations. More results and visualizations of
the weights assigned to suboptimal demonstrations are provided in
the supplementary material.

\section{Conclusion}

We propose ILMAR, an imitation learning method designed for datasets
with a limited number of expert demonstrations and supplementary
suboptimal demonstrations. By utilizing a functional of the
advantage function, ILMAR avoids directly discarding high-quality
non-expert demonstrations in the supplementary suboptimal dataset,
thereby improving the utilization of the supplementary
demonstrations. To further enhance the updating of our
discriminator, we introduce the meta-goal method, leading to
performance improvements. Experimental results show that ILMAR
achieves performance that is competitive with or superior to
state-of-the-art IL algorithms. One potential direction for future
exploration is to investigate the use of a functional of the
advantage function weighting for processing suboptimal
demonstrations with reward information. Another direction is to
apply the meta-goal method to semi-supervised learning scenarios.

\bibliographystyle{named}
\bibliography{ijcai25}

\clearpage

\appendix

\section{Experimental Details}

\subsection{Implementation Detail}

In this section, we provide detailed descriptions of our
experimental setup to ensure the reproducibility of our results. We
evaluate the performance of various imitation learning algorithms
across four motion control tasks in the MuJoCo suite
\cite{DBLP:conf/iros/TodorovET12}: Ant-v2, Humanoid-v2, Hopper-v2,
and Walker2d-v2.

\begin{figure}[t]
    \centering
        \includegraphics[width=0.9\linewidth]{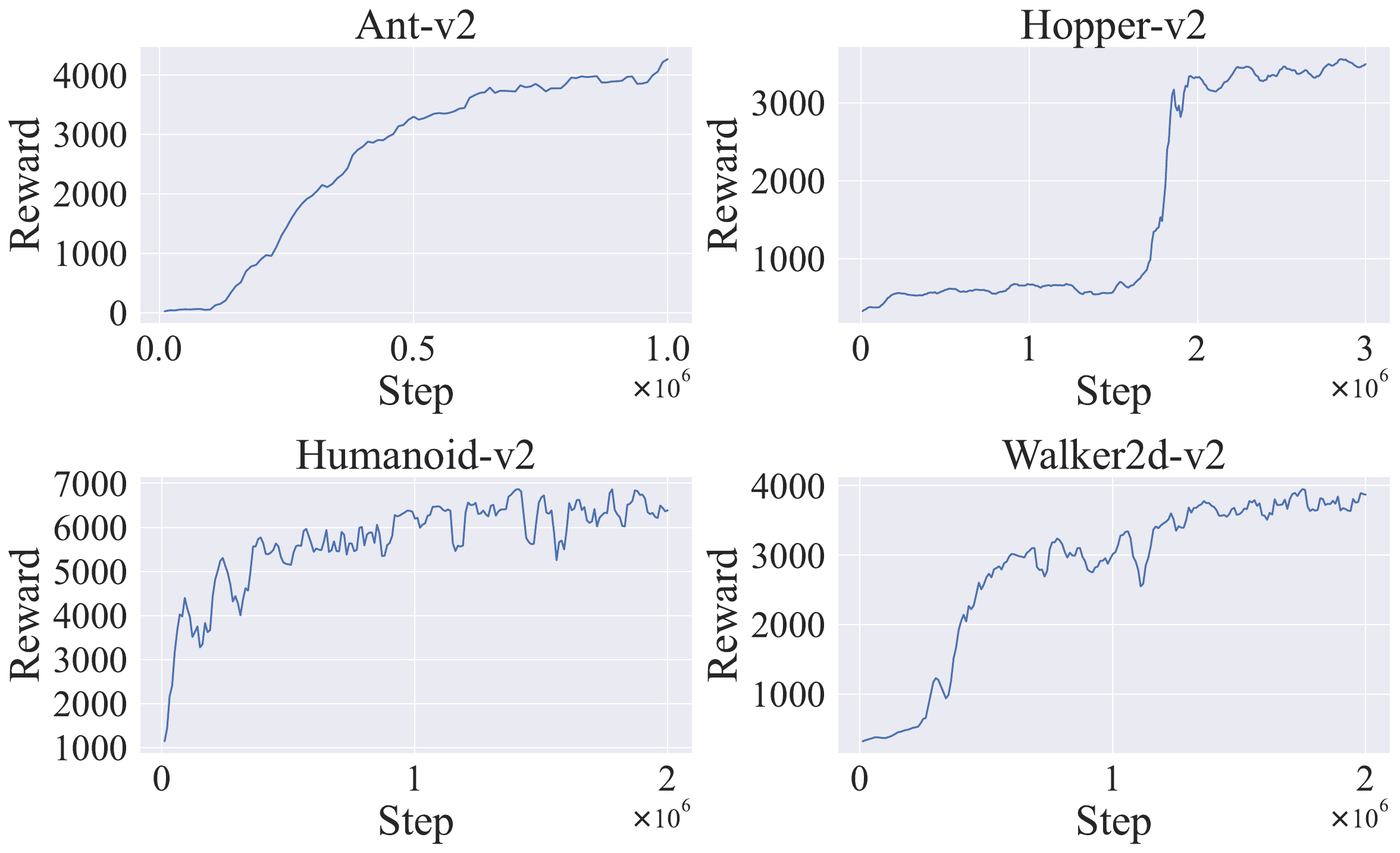}
        \caption{Training curves of expert agents on 4 locomotion control environments.}
        \label{fig:expert}
\end{figure}

To train the expert policies, we use the proximal policy
optimization (PPO) \cite{DBLP:journals/corr/SchulmanWDRK17} and soft
actor-critic (SAC) \cite{DBLP:journals/corr/abs-1812-05905}
algorithms. After comparing the results, we select the
best-performing models as the expert policies: Ant-v2 is trained
with PPO for one million steps, Humanoid-v2 with SAC for two million
steps, Hopper-v2 with PPO for three million steps, and Walker2d-v2
with PPO for two million steps. We select four intermediate policies
with varying levels of optimality by evaluating the policies every
100,000 steps, using the best-performing policy as the expert
policy. All algorithmic dependencies in our code are based on the
implementation provided by CAIL \cite{DBLP:conf/nips/ZhangCSS21},
available at
\url{https://github.com/Stanford-ILIAD/Confidence-Aware-Imitation-Learning}.
The training curves of the expert agents are presented in
Figure~\ref{fig:expert}.

We select four suboptimal policies, each with performance rewards
evaluated every 10,000 steps with 5 episodes, approximating 80\%,
60\%, 40\%, and 20\% of the optimal policy, respectively.
Table~\ref{tab:demon} illustrates the average performance of the
collected trajectories.

\begin{table}[t]
    \centering
    \resizebox{0.68\linewidth}{!}{
    \begin{tabular}{l|cccc}
        \toprule
        Environment & Ant & Hopper & Humanoid & Walker2d \\
        \midrule
        Expert    & 4761     & 3635   & 7025  & 4021 \\
        Police 1  & 3762     & 3329   & 5741  & 3218 \\
        Police 2  & 2879     & 3301   & 5084  & 2897 \\
        Police 3  & 2161     & 2593   & 4334  & 1305\\
        Police 4  & 796      & 637    & 2683  & 744\\
        \bottomrule
    \end{tabular}}
    \caption{The performance of policies used in our experimental results.}
    \label{tab:demon}
\end{table}

All three task settings share the same expert dataset $D^E$, which
consists of only one expert trajectory. The supplementary dataset
for each task is composed of a mixture of expert and suboptimal
trajectories at different ratios: 1:0.25 (T1), 1:1 (T2), and 1:4
(T3). Specifically, the supplementary suboptimal dataset $D^S$
includes 400 expert trajectories combined with 100, 400, and 1600
suboptimal trajectories generated by the four intermediate policies,
where the number of trajectories contributed by each suboptimal
policy is equal.

We visualize the distributional differences between the additional
datasets and the expert demonstrations under three task settings on
Ant-v2. Specifically, we train a variational autoencoder (VAE)
\cite{DBLP:journals/corr/KingmaW13} to reconstruct state-action
pairs. To highlight the differences in their underlying
distributions, we visualize the log-likelihoods of the additional
datasets under various task settings using kernel density estimation
(KDE) plots. As shown in Figure~\ref{fig:task_setting}, when the
proportion of expert demonstrations in the supplementary dataset is
high, the distributional differences between the expert and
supplementary datasets are relatively small. However, as the
proportion of expert demonstrations decreases, the distributional
differences gradually increase, posing greater challenges to the
algorithm in handling suboptimal demonstrations.

\begin{table}[t]
    \centering
    \resizebox{1.0\linewidth}{!}{
    \begin{tabular}{c|cccccc}
        \toprule
        Hyperparameters & BC & ISW-BC & DemoDICE & ILMAR & BCND & DWBC\\
        \midrule
        Learning rate (actor)   & $3\times 10^{-4}$     & $3\times 10^{-4}$     & $3\times 10^{-4} $    & $3\times 10^{-4}$     &$3\times 10^{-4} $ &$1\times 10^{-4} $\\
        Network size (actor)    & $[256,256]$           & $[256,256]$           & $[256,256] $          & $[256,256]$   & $[256,256]$   & $[256,256]$ \\
        Learning rate (critic)    & -   & -  & $3\times 10^{-4}$ & - & -  & -\\
        Network size (critic)  & -  & -   & $[256,256] $ & - & - & - \\
        Learning rate (discriminator)  & -    & $3\times 10^{-4}$  & $3\times 10^{-4}$  &$3\times 10^{-4}$ & - &$1\times 10^{-4} $\\
        Network size (discriminator)  & -  & $[256,256] $   & $[256,256] $ & $[256,256]$ &- &$[256,256]$\\
        Batch size   &256 &   256      & 256  &256 &256 &256 \\
        \midrule
        Training iterations        &$1 \times 10^{6}$ &   $1 \times 10^{6}$      & $1 \times 10^{6}$  &$1 \times 10^{6}$  &$1 \times 10^{6}$  &$1 \times 10^{6}$ \\
        \bottomrule
    \end{tabular}}
    \caption{Configurations of hyperparameters used in our experiments.}
    \label{tab:hyp}
\end{table}

\begin{figure}[t]
    \centering
    \includegraphics[width=0.65\columnwidth]{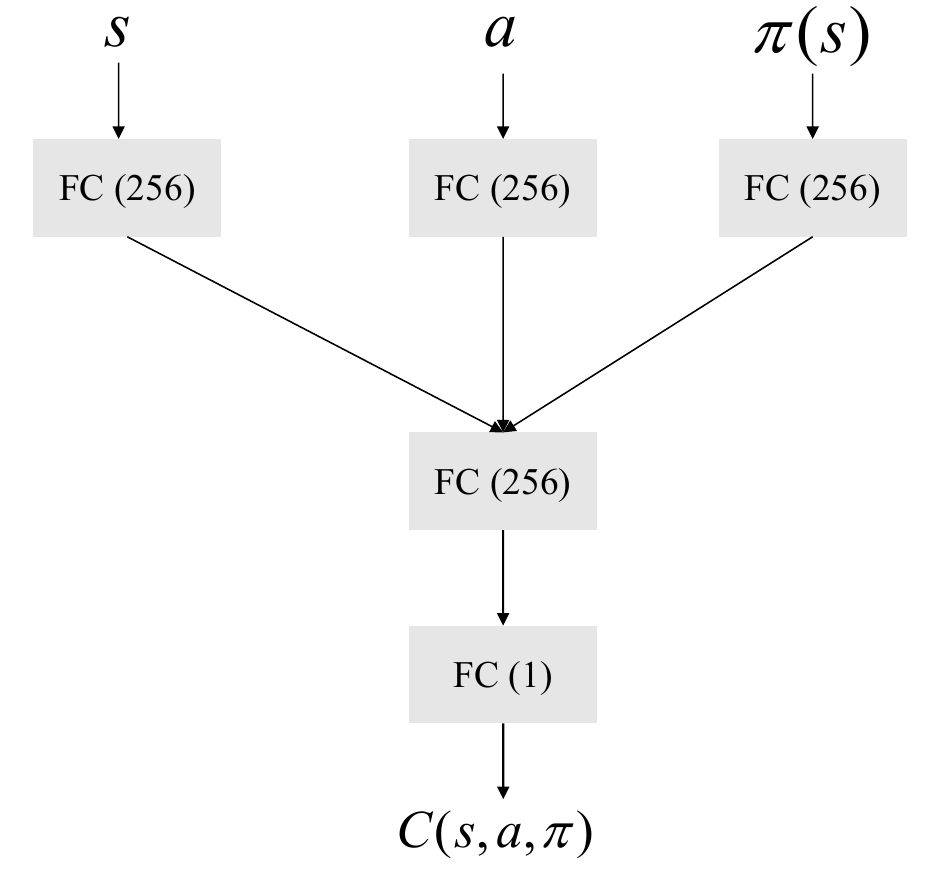}
    \caption{Structure of the discriminator network $C$. Action $a$ and $\pi(s)$
are independently input into the same fully connected (FC) neural
network (action encoder).}
    \label{fig:structure}
\end{figure}

We report the mean and standard error of performance across five
different random seeds (2023, 2024, 2025, 2026, 2027). The
experiments were conducted using GTX 3090 GPUs, Intel Xeon Silver
4214R CPUs, and Ubuntu 20.04 as the operating system. The DemoDICE
codebase is based on the original work by the authors, available at
\url{https://github.com/KAIST-AILab/imitation-dice}, the DWBC
codebase can be accessed at \url{https://github.com/ryanxhr/DWBC}
and the ISW-BC codebase can be accessed at
\url{https://github.com/liziniu/ISWBC}. Since the official
implementation of BCND is not publicly available, we reproduce its
method based on the description in its paper. We set its
hyperparameters to $M$=10 and $K$=1 To ensure stable discriminator
learning, and gradient penalty regularization is applied during
training as proposed in \cite{DBLP:conf/nips/GulrajaniAADC17}, to
enforce the 1-Lipschitz constraint. Detailed hyperparameter
configurations used in our main experiments are provided in
Table~\ref{tab:hyp}. In our approach, fully connected neural (FC)
networks with ReLU activations are used for all function
approximators. For the policy networks, we adopt a stochastic policy
(Gaussian), where the model outputs the mean and variance of the
action using the Tanh function. The Adam optimizer is selected for
the optimization process across all models. It is worth noting that
the discriminators for DWBC and ILMAR do not take $(s,a)$ directly
as input. The specific architecture of DWBC is described in its
original paper, while the architecture of our model is illustrated
in Figure~\ref{fig:structure}. For the ILMAR-specific
hyperparameters $\alpha$ and $\beta$, they both are set to 1 for
tasks T1 and T2. In T3, a grid search is performed over $\alpha$ and
$\beta$ to better understand the roles of the vanilla loss and meta
loss. The optimal hyperparameter settings are determined through
this search, as detailed in Table~\ref{tab:spe-hyp}.

\begin{table}[t]
    \centering
    \resizebox{0.8\linewidth}{!}{
    \begin{tabular}{c|cccc}
        \toprule
        Hyperparameters & Ant & Hopper & Humanoid & Walker2d\\
        \midrule
        $\alpha$   & $0.3$     & $0.05$     & $0.05 $    & $0.7$    \\
        $\beta$    & $0.01$           & $0.01$           & $0.15$          & $0.01$    \\
        \bottomrule
    \end{tabular}}
    \caption{Specific hyperparameters used in ILMAR.}
    \label{tab:spe-hyp}
\end{table}

\subsection{Additional Experimental Results}

\subsubsection{\texorpdfstring{Ablation Test on Hyperparameters $\alpha$ and $\beta$}{Ablation Test on Hyperparameters Alpha and Beta}}

As we can notice in Eq.~\eqref{eq:total_loss}, ILMAR introduces
hyperparmeters $\alpha$ and $\beta$, which control the relative
influence of the meta loss and the vanilla loss on the updates of
the discriminator. In this section, we aim to study how does the
choice of $\alpha$ and $\beta$ affect the training processes and
performance of our algorithm. We conduct a grid search over $\alpha
\in \{ 0, 0.1,0.3, 0.7, 1.0\}$ and $\beta \in \{ 0, 0.01,0.05, 0.5,
1.0\}$ in the task setting T3.

\begin{figure}[t]
    \centering
    \includegraphics[width=1.0\columnwidth]{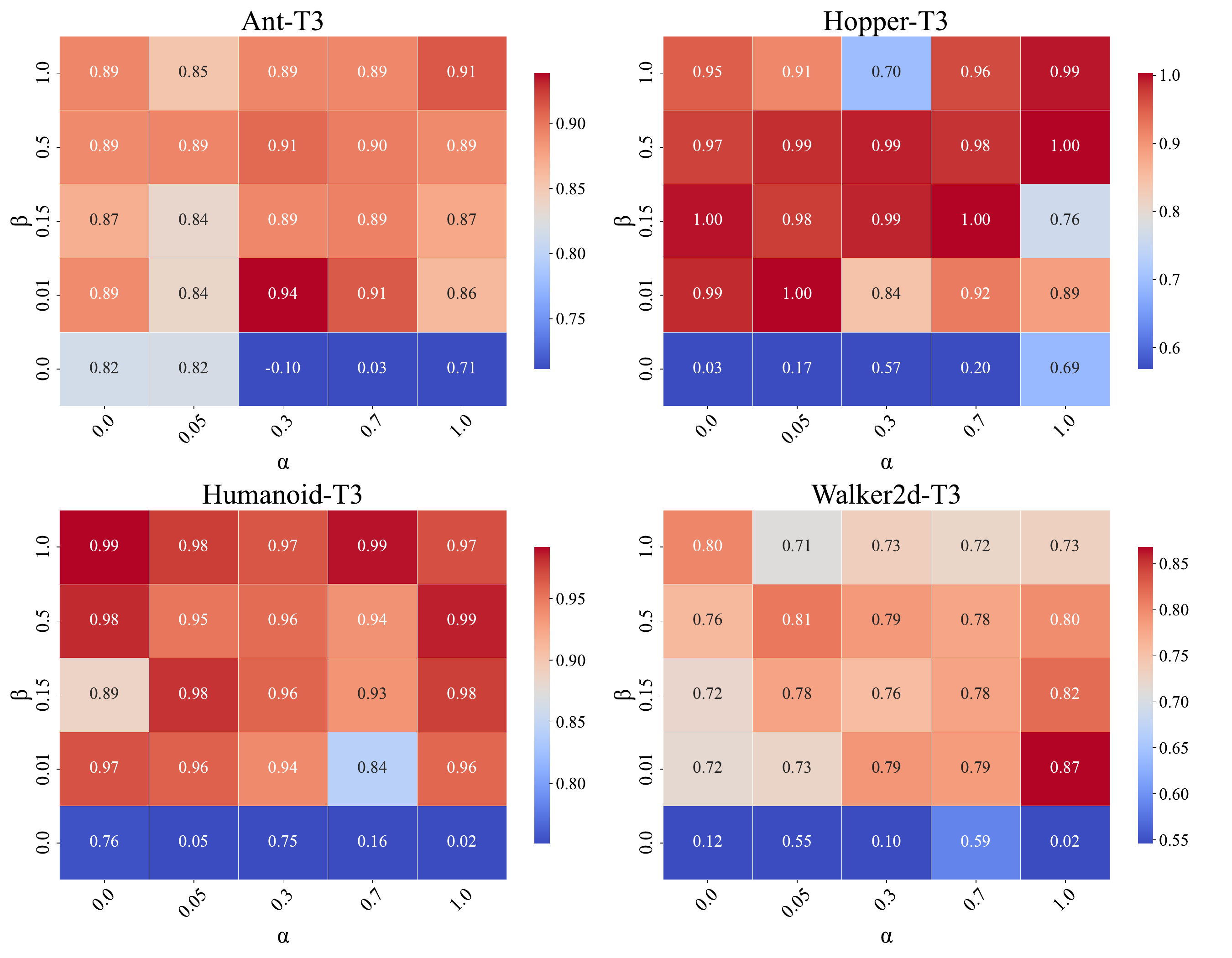}
    \caption{The normalized scores in the task setting T3 with seed 2025, obtained by
sweeping through combinations of meta loss strength $\alpha$ and
vanilla loss strength $\beta$.}
    \label{fig:heatmap_all}
\end{figure}

\begin{figure}[t]
\centering
\includegraphics[width=1.0\linewidth]{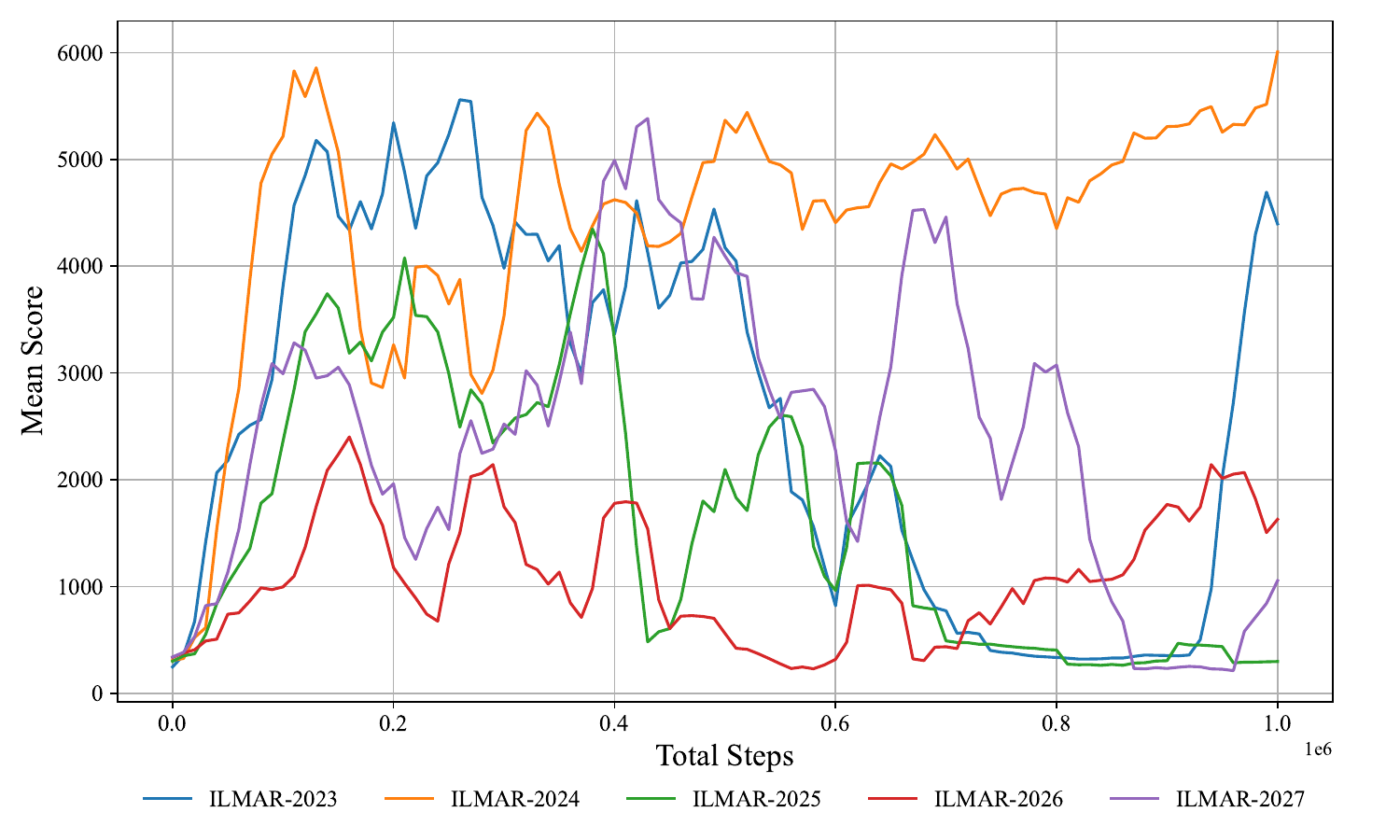}
\caption{Training curves of ILMAR without the vanilla loss with five
seeds on Humanoid-v2 in the task setting T3.}
\label{fig:meta_goal_Humnaoid}
\end{figure}

Figure~\ref{fig:heatmap_all} demonstrates that meta-goal
significantly enhances policy performance. We observe that when
using the meta-goal method alone, ILMAR performs suboptimally.
However, when combined with the vanilla loss, the performance of
ILMAR is often significantly improved. Notably, in scenarios where
only meta-goal is used, the performance exhibits higher sensitivity
to random seeds. To illustrate this, we present the performance
curves under the setting $\alpha=1, \beta=0$ across five random
seeds, as shown in Figure~\ref{fig:meta_goal_Humnaoid}. Meanwhile,
we observe that when both the vanilla loss and the meta loss are
used to update the discriminator, ILMAR demonstrates robustness to
the choice of $\alpha$ and $\beta$, maintaining stable performance
across nearly all hyperparameter configurations.

This observation supports our previous theoretical analysis, which
suggests that the explicitly designed vanilla loss provides prior
knowledge by regularizing the output of the discriminator. This
prior knowledge assists the model in learning how to effectively
weight the demonstrations using meta-goal, ultimately yielding a
policy that closely resembles the expert policy.

\subsubsection{Additional Analysis of Results}

\begin{figure}
\centering
\includegraphics[width=0.9\linewidth]{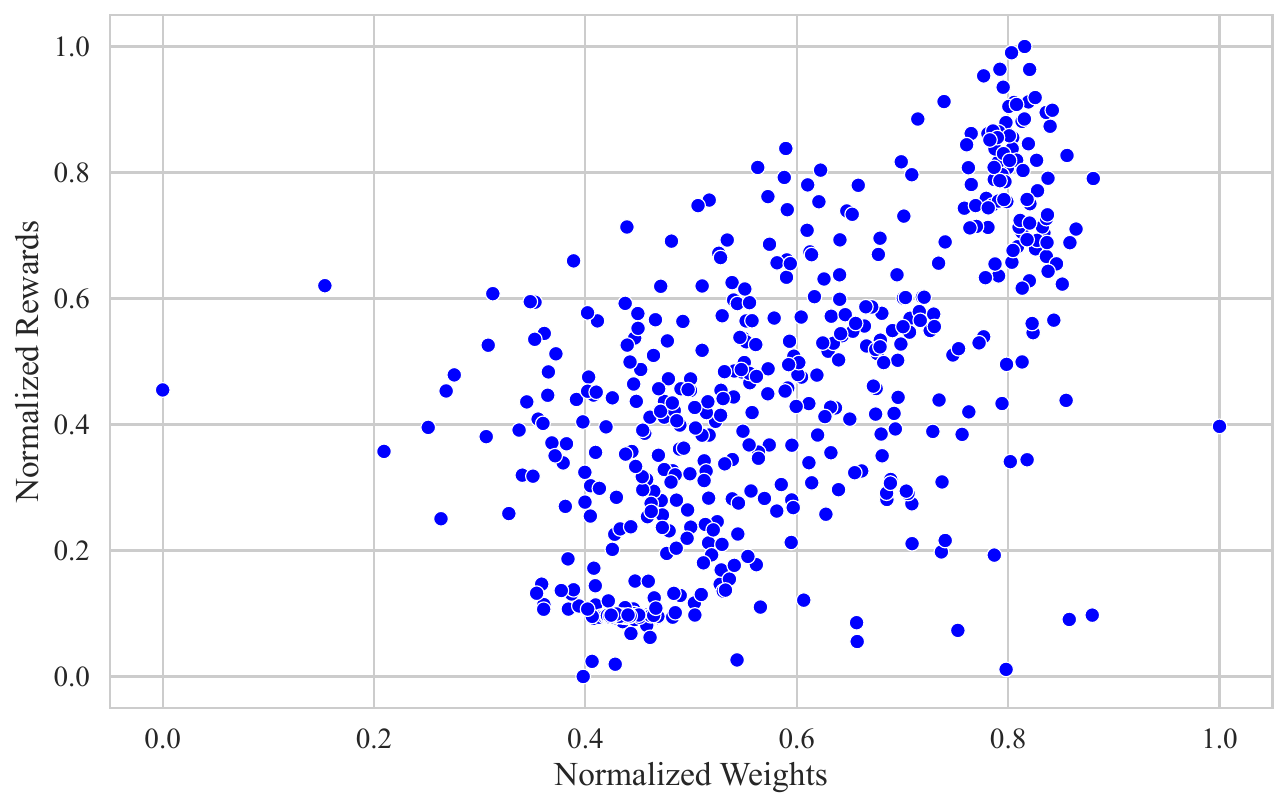}
\caption{The actual reward and the weights assigned by ILMAR for
suboptimal demonstrations in Ant-v2.} \label{fig:rewards_weights}
\end{figure}

Figure~\ref{fig:rewards_weights} illustrates the relationship
between min-max normalized weights and rewards in the Ant-v2 task
setting T3. From the figure, it is evident that the weights assigned
by ILMAR exhibit a clear monotonic positive correlation with the
true rewards.

\subsubsection{Additional Experiments of More Expert Demonstrations}

\begin{figure}[t]
    \centering
    \includegraphics[width=1.0\linewidth]{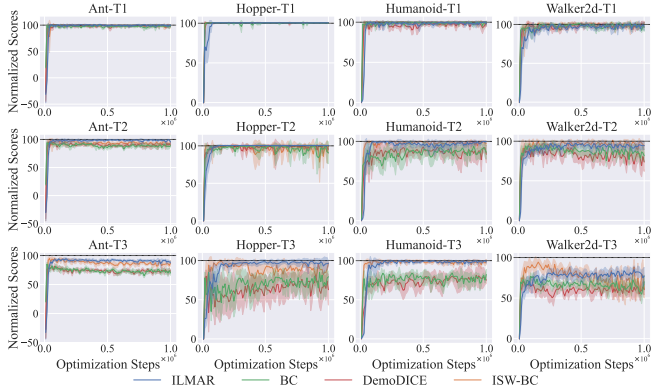}
    \caption{Training curves of ILMAR and baseline algorithms on tasks T1, T2,
T3. The y-axis represents the normalized scores of the algorithm
during training. The solid line corresponds to the average
performance under five random seeds, and the shaded area corresponds
to the 95\% confidence interval.}
    \label{fig:baseline_5}
\end{figure}

Furthermore, to investigate the performance of ILMAR with a larger
amount of expert data, we conduct experiments by increasing the
number of expert demonstrations in expert dataset to 5. We set both
$\alpha$ and $\beta$ to 1 to minimize the impact of hyperparameters
selection on the results. The results, as shown in
Figure~\ref{fig:baseline_5} and Table~\ref{tab:5 expert
demonstrations}, demonstrate that ILMAR continues to exhibit a
significant advantage over other algorithms even with an increased
number of expert demonstrations.

\begin{table}[t]
    \centering
    \resizebox{1.0\linewidth}{!}{
    \begin{tabular}{ccccccc}
        \toprule
        Task setting & Environment & Ant & Hopper & Humanoid & Walker2d  & Score \\
        \midrule
        & Random  & -75  & 15   & 122 & 1   & 0\\
        & Expert  & 4761  & 3635   & 7025 & 4021   & 100\\
        \midrule
        \multirow{6}{*}{\textbf{T1}}
        & BC  & 4649$_{\pm 69}$    & $3650{\pm 3}$  & $6959{\pm 72}$  & $3882{\pm 92}$ & 98.42\\
        & DemoDICE   &$4658_{\pm 56}$ &   $3651_{\pm 1}$      & $6761_{\pm 237}$  &$3988_{\pm 39}$ & 98.42\\
        & ISW-BC    &$4649_{\pm 83}$  &   $3652_{\pm 2}$  &$6977_{\pm 83}$ &$\boldsymbol{4007_{\pm 63}}$ & 99.29  \\
        & ILMAR (ours) &$\boldsymbol{4728_{\pm 27}}$   &$\boldsymbol{3653_{\pm 1}}$       & $\boldsymbol{6997_{\pm {53}}}$  & $3937_{\pm83}$ & $\boldsymbol{99.33}$  \\
        \midrule
        \multirow{6}{*}{\textbf{T2}}
        & BC  & 4222$_{\pm82}$    & $3461_{\pm 335}$  & $6266_{\pm 345}$  & $3378_{\pm 108}$ & 89.26\\
        & DemoDICE   &$4294_{\pm 76}$ &   $3533_{\pm 116}$      & $6127_{\pm 583}$  &$3145_{\pm 474}$ & 88.18\\
        & ISW-BC    &$4459_{\pm 69}$  &   $3467_{\pm 282}$  &$\boldsymbol{6916_{\pm 124}}$ &$\boldsymbol{3145_{\pm 474}}$ & 95.90  \\
        & ILMAR (ours) &$\boldsymbol{4643_{\pm 54}}$   &$\boldsymbol{3648_{\pm 6}}$       & $6980_{\pm 48}$  & $3806_{\pm {95}}$ & $\boldsymbol{97.98}$  \\
        \midrule
        \multirow{6}{*}{\textbf{T3}}
        & BC  & 3411$_{\pm166}$    & $2704_{\pm 388}$  & $5420_{\pm 205}$  & $2454_{\pm 267}$ & 71.04\\
        & DemoDICE   &$3457_{\pm 50}$ &   $2442_{\pm 566}$      & $5521_{\pm 715}$  &$2417_{\pm 257}$ & 69.60\\
        & ISW-BC    &$4046_{\pm 139}$  &   $3192_{\pm 369}$  &$6860_{\pm 129}$ &$2775_{\pm 376}$ & 84.90  \\
        & ILMAR (ours) &$\boldsymbol{4272_{\pm 139}}$   &$\boldsymbol{3472_{\pm 303}}$       & $\boldsymbol{6945_{\pm {32}}}$  &$\boldsymbol{3196_{\pm {239}}}$ & $\boldsymbol{90.93}$  \\
        \bottomrule
    \end{tabular}}
    \caption{Performance of ILMAR and baseline algorithms on Ant-v2, Hopper-v2, Walker2d-v2 and Humanoid-v2 over the final 5 evaluations and 5 seeds.
    The best results are in bold.
    ILMAR significantly outperforms existing imitation learning methods from suboptimal datasets.}
    \label{tab:5 expert demonstrations}
\end{table}

\begin{table}[t]
    \centering
    \resizebox{1.0\linewidth}{!}{
    \begin{tabular}{cc|cccc|c}
        \toprule
        Environment && Ant & Hopper & Humanoid & Walker2d  & Score \\
        \midrule
        \multirow{2}{*}{DemoDICE}   &&$3457_{\pm 50}$ &   $2442_{\pm 566}$      & $5521_{\pm 715}$  &$2417_{\pm 257}$ & 69.60\\
        &+Meta-goal&$\boldsymbol{3815_{\pm 61}}$  &   $\boldsymbol{2655_{\pm 603}}$  &$\boldsymbol{5755_{\pm 373}}$ &$\boldsymbol{2699_{\pm 188}}$ & $\boldsymbol{75.52}$  \\
        \midrule
        \multirow{2}{*}{ISW-BC}   &&$4046_{\pm 139}$  &   $3192_{\pm 369}$  &$6860_{\pm 129}$ &$2775_{\pm 376}$ & 84.90\\
         &+Meta-goal    &$\boldsymbol{4216{\pm 124}}$  &   $3078_{\pm 221}$  &$\boldsymbol{6955_{\pm 52}}$ &$\boldsymbol{3531_{\pm 267}}$ & $\boldsymbol{90.03}$  \\
        \bottomrule
    \end{tabular}}
     \caption{Performance of other algorithms using meta-goal on the MuJoCo
environments when the expert dataset includes five expert
trajectories. The best results are in bold.}
    \label{tab:meta-goal_5}
\end{table}

We apply meta-goal to DemoDICE and ISW-BC when the expert dataset
includes five expert trajectories. The results are presented in
Table~\ref{tab:meta-goal_5}.

\begin{figure}[!t]
    \centering
    \includegraphics[width=0.75\columnwidth]{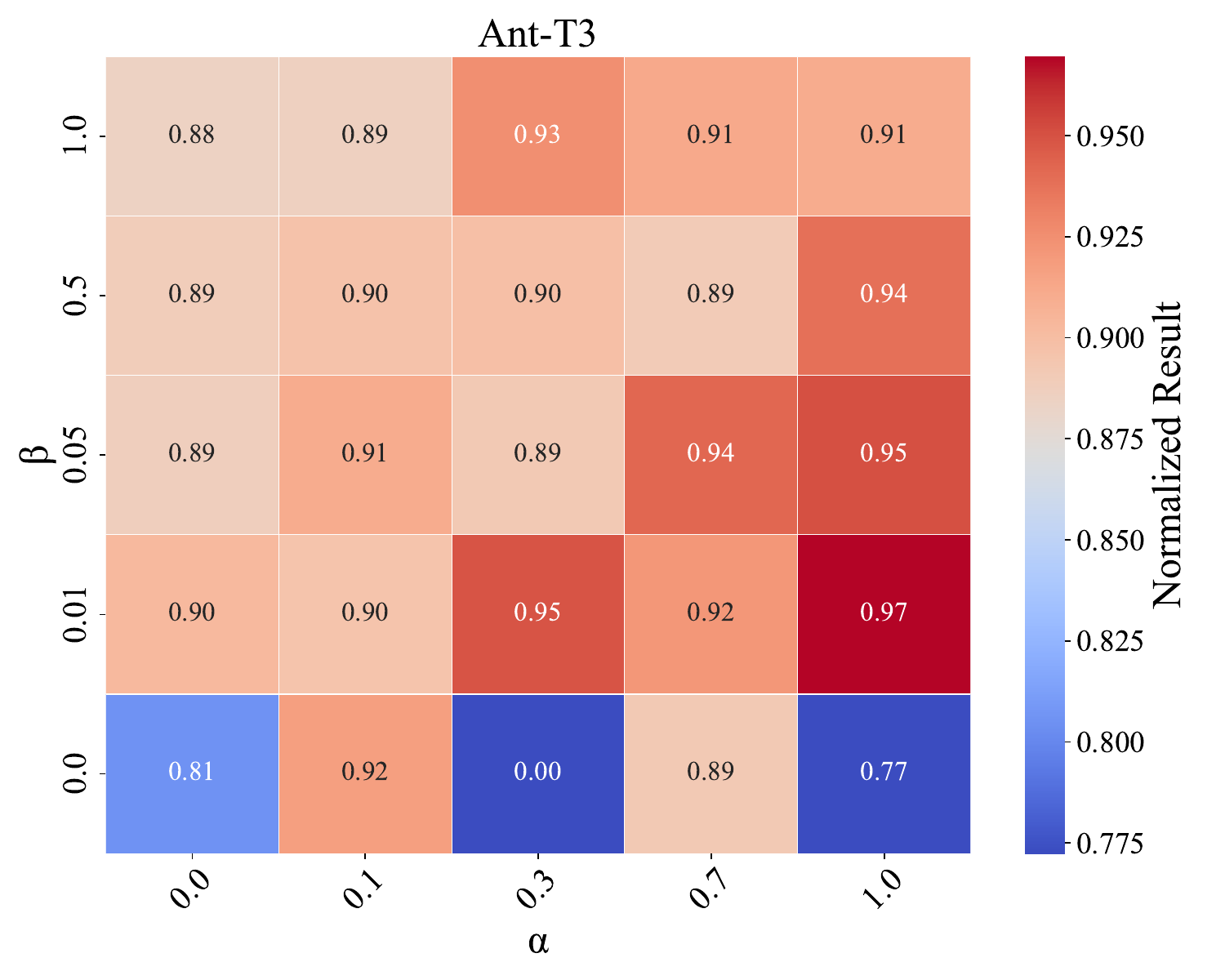}
    \caption{The normalized scores on Ant-v2 in the task setting T3 with seed 2025,
obtained by sweeping through combinations of meta loss strength
$\alpha$ and vanilla loss strength $\beta$ when the expert dataset
includes five expert trajectories.}
    \label{fig:heatmap}
\end{figure}

We conduct a grid search over $\alpha \in \{ 0, 0.1,0.3, 0.7, 1.0\}$
and $\beta \in \{ 0, 0.01,0.05, 0.5, 1.0\}$ on Ant-v2 in the task
setting T3. The results are presented in Figure~\ref{fig:heatmap}.

\subsubsection{Convergence Analysis}

\begin{figure}[t]
    \centering
    \includegraphics[width=1.0\linewidth]{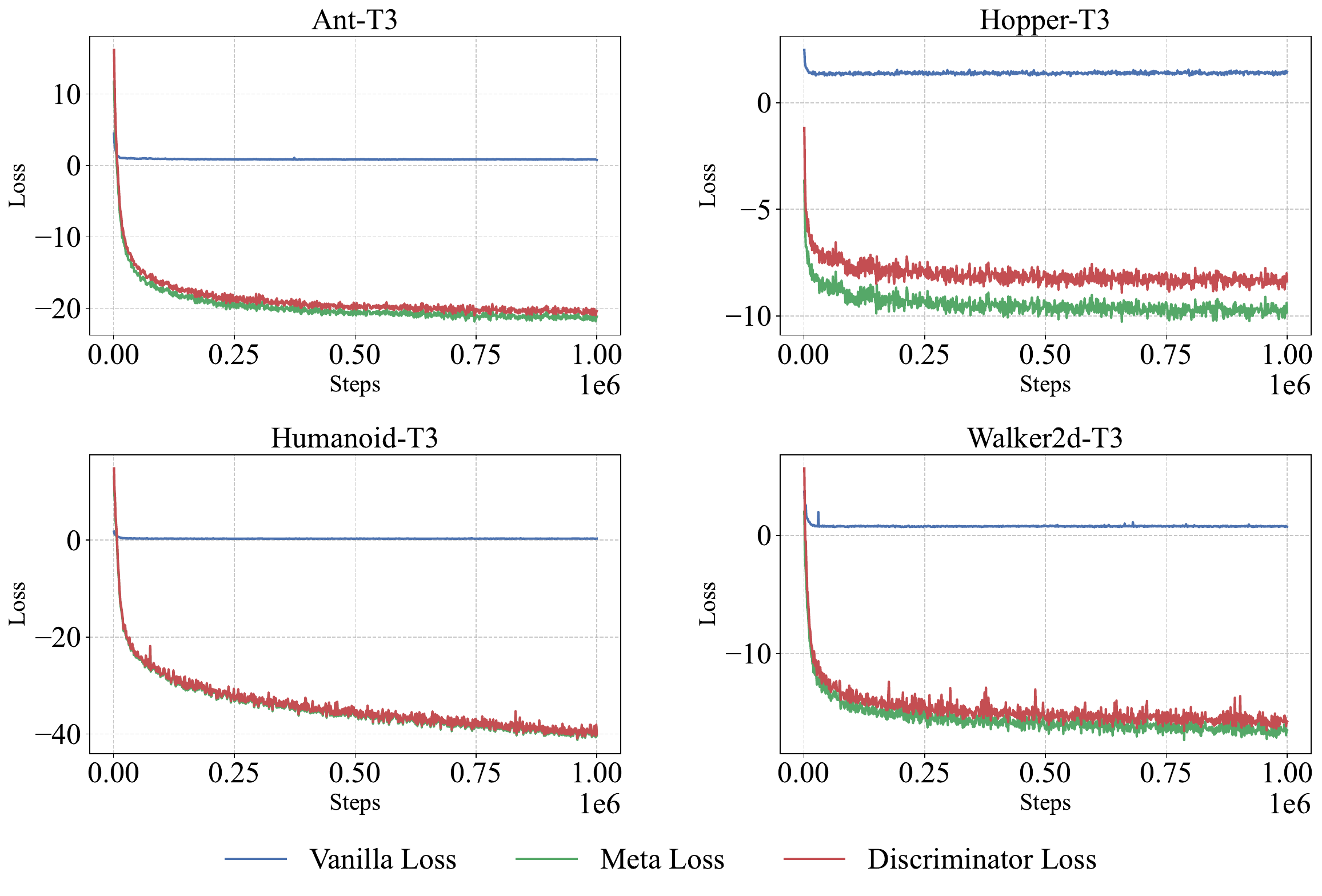}
    \caption{Loss curves of ILMAR on the MuJoCo environments in the task setting T3.}
    \label{fig:loss}
\end{figure}

\begin{figure}[t]
    \centering
    \includegraphics[width=1.0\linewidth]{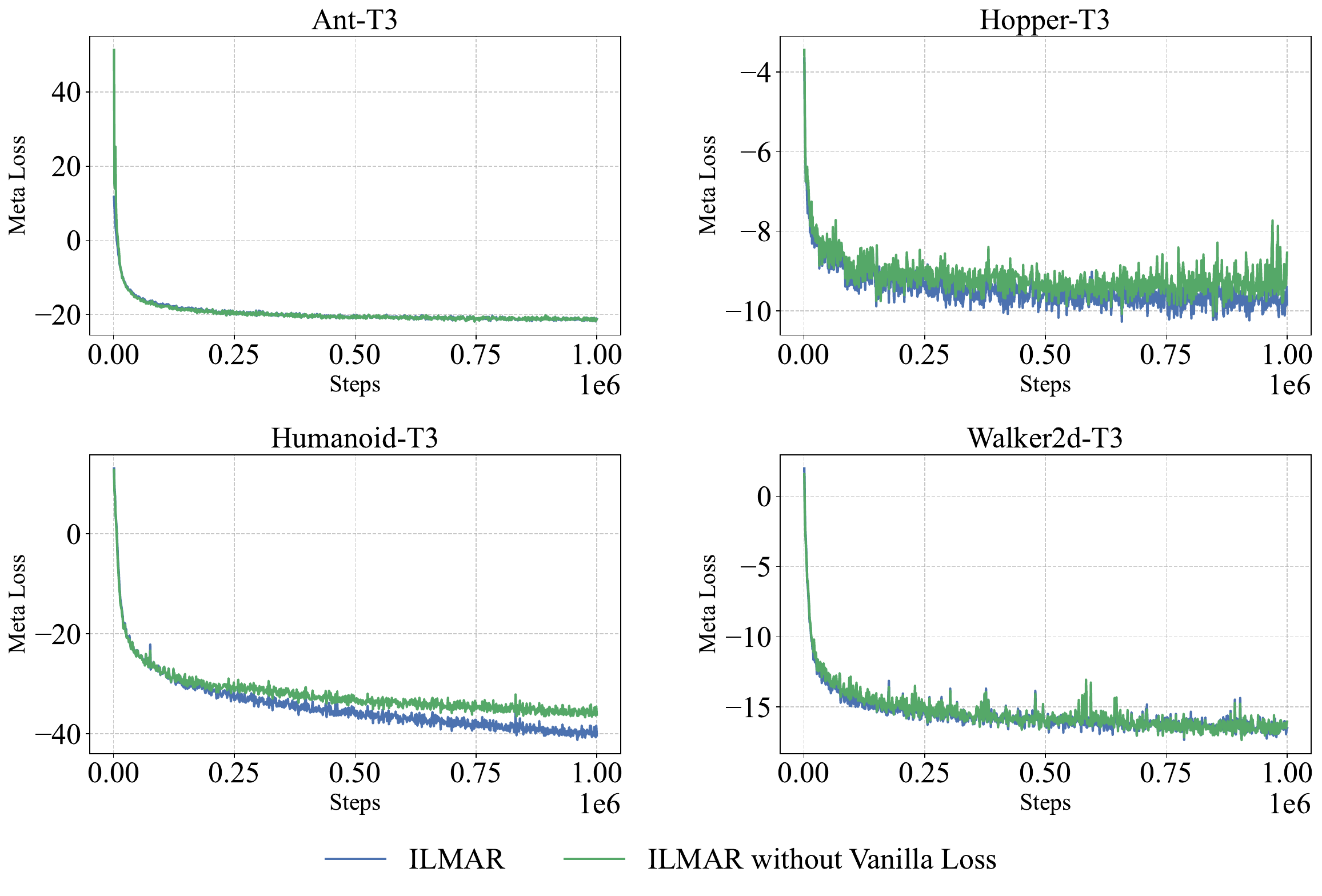}
    \caption{Meta loss curves of ILMAR on the MuJoCo environments in the task
setting T3. The green line represents updating the discriminator
using only the meta-goal, while the blue line represents updating
the discriminator with both the vanilla loss and meta-loss.}
    \label{fig:meta loss compare}
\end{figure}

To better understand why we recommend combining meta-goal with the
original weighted behavior cloning instead of using it as a
standalone algorithm, we analyze the loss under different scenarios.
In this section, we set both $\alpha$ and $\beta$ to 1 to minimize
the impact of hyperparameter selection on the results.
Figure~\ref{fig:loss} illustrates the variation in the discriminator
loss during training on T3. As shown in Figure~\ref{fig:loss},
vanilla loss converges rapidly in the early stages of training,
after which the updates to the discriminator are primarily
influenced by the meta loss. As training progresses, the meta loss
gradually decreases and converges. We also compare the loss dynamics
when the discriminator is updated using only meta-goal.
Figure~\ref{fig:meta loss compare} illustrates the variation in the
meta loss of ILMAR when trained without the vanilla loss
discriminator on task T3, i.e., with $\alpha=1, \beta=0$. From the
figure, we observe that incorporating the vanilla loss not only does
not hinder the convergence of the meta-loss but actually helps it
converge to a lower value. This aligns with our theoretical
analysis, which suggests that the vanilla loss provides the
discriminator with prior knowledge, aiding the convergence of the
meta-loss and thereby improving the overall performance of the
policy.

These observations validate the theoretical insights: the design of
the vanilla loss provides prior knowledge that constrains the
direction of updates of discriminator, facilitating the convergence
of the meta loss and ultimately enhancing the performance of model.

\section{Theoretical Derivation}

\subsection{Derivation of The Gradient \texorpdfstring{$\frac{\partial L_{meta}}{\partial \psi}$}{dL_meta/dpsi}}

Eq.~\ref{eq:chain rule} establishes that the gradient
$\frac{\partial L_{meta}}{\partial \psi}$ can be expressed as:
\[
     \mu \frac{1}{|\mathcal D|} \frac{\partial L_{meta}}{\partial \theta_{t+1}}  \sum _{(s,a)\in \mathcal D} \frac {\partial^2 w(s,a,\pi_{\theta_t}) \log \pi_{\theta_t}(a|s)}{\partial \psi \partial \theta_t}.
\]
Here, we provide the detailed derivation process for this result.
\begin{align}
    &\frac{\partial L_{meta}}{\partial \psi}=\frac{\partial L_{meta}}{\partial \theta_{t+1}} \frac{\partial \theta_{t+1}}{\partial \psi} \nonumber \\
    &=\frac{\partial L_{meta}}{\partial \theta_{t+1}} \frac{\partial {\theta_t - \mu \nabla_\theta L_{actor}(s,a;\theta_t,\psi_{t})}}
    {\partial \psi} \nonumber \\
    &=-\mu \frac{\partial L_{meta}}{\partial \theta_{t+1}} \frac{\partial { \frac{\partial L_{actor}}{\partial \theta_t} (s,a;\theta_t,\psi_{t})}}    {\partial \psi} \nonumber \\
    &=-\mu \frac{\partial L_{meta}}{\partial \theta_{t+1}} \frac {\partial^2 L_{actor}}{\partial \psi \partial \theta_t} \nonumber \\
    &= -\mu \frac{\partial L_{meta}}{\partial \theta_{t+1}} \frac {\partial^2 - \frac{1}{|\mathcal D|}\sum _{(s,a)\in \mathcal D}w(s,a,\pi_{\theta_t}) \log \pi_{\theta_t}(a|s)}{\partial \psi \partial \theta_t} \nonumber\\
    &= \mu \frac{1}{|\mathcal D|} \frac{\partial L_{meta}}{\partial \theta_{t+1}}  \sum _{(s,a)\in \mathcal D} \frac {\partial^2 w(s,a,\pi_{\theta_t}) \log \pi_{\theta_t}(a|s)}{\partial \psi \partial \theta_t} \nonumber
\end{align}

\subsection{Proof of Theorem~\ref{th:convergence}}

\begin{proof}
By Lemma 2 in \citet{DBLP:conf/nips/ZhangCSS21} (with proof on Page
12), the function $f(x)$ is Lipschitz-smooth with constant $L$, then
the following inequality holds:
\[
    f(y) \leq f(x)+\nabla f(x)^{T}(y-x)+\frac{L}{2}\|y-x\|^{2}, \quad \forall x,
    y.
\]
Thus, we have:
\begin{proof}
\begin{align}
&\mathcal{L}_{C}\left(\theta_{t+1}\right)-\mathcal{L}_{C}\left(\theta_{t}\right)  \nonumber \\
&\leq \nabla_{\theta} \mathcal{L}_{C}\left(\theta_{t}\right)^{T}\left(\theta_{t+1}-\theta_{t}\right)+\frac{L}{2}\left\|\left(\theta_{t+1}-\theta_{t}\right)\right\|^{2} \\ &= -\mu \nabla_{\theta} \mathcal{L}_{C}\left(\theta_{t+1}\right)^{T} \nabla_{\theta} \mathcal{L}_{\text {actor }}\left(\theta_{t}, \psi_{t}\right) \nonumber \\
&+\frac{L}{2} \mu^{2}\left\|\nabla_{\theta} \mathcal{L}_{\text {actor }}\left(\theta_{t}, \psi_{t}\right)\right\|^{2} \nonumber\\
&\leq-\left(\mu K-\frac{L}{2} \mu^{2}\right)\left\|\nabla_{\theta} \mathcal{L}_{\text {actor }}\left(\theta_{t}, \psi_{t}\right)\right\|^{2} \\
&\leq 0
\end{align}
\end{proof}
The first inequality comes from Lemma 2 in
\citet{DBLP:conf/nips/ZhangCSS21}, and the second inequality comes
from that only when $\nabla_\theta\mathcal{L}_{C}(\theta_{t+1}
)^\top\nabla_\theta\mathcal{L}_{actor}(\theta_t,\psi_{t})\geq
K||\nabla_\theta\mathcal{L}_{actor}(\theta_t,\psi_{t})||^2$ holds,
we update the policy and the learning rate satisfies $\mu \leq
\frac{2K}{L}$.
\end{proof}

\end{document}